%% file: main.tex
\documentclass[]{primo}
\usepackage{fix-cm}
\usepackage{csquotes}
\usepackage{afterpage}
\usepackage{mathtools}
\usepackage{mathrsfs}
\usepackage{xspace}
\usepackage{colortbl}
\usepackage{enumitem}
\usepackage{wrapfig}
\usepackage{makecell}
\usepackage{pifont}
\usepackage{array}
\usepackage{tabularx}
\usepackage{longtable}
\usepackage{xltabular}
\usepackage{float}
\usepackage{placeins}
\usepackage[title]{appendix}

\graphicspath{{figures/}{assets/}}


\definecolor{fbApp}{HTML}{c8e7fa}
\definecolor{fbPurple3}{HTML}{f0ebf5}
\definecolor{yuhengBg}{HTML}{fdf0ed}
\definecolor{yuhengAccent}{HTML}{ce553b}
\definecolor{citecolor}{HTML}{0071BC}
\definecolor{linkcolor}{HTML}{ED1C24}
\definecolor{emcolor}{HTML}{000000}
\newcommand{\myemcolor}{emcolor}

\addto\captionsenglish{}
\titleformat*{\paragraph}{\rmfamily\bfseries}

\newcommand{\method}{\textbf{Yuheng}\xspace}
\newcommand{\methodfour}{\textbf{Yuheng-4B}\xspace}
\newcommand{\methodtwentyseven}{\textbf{Yuheng-27B}\xspace}
\newcommand{\bench}{\textbf{Yuheng Bench}\xspace}

\newcommand{\anchormech}{in-context reference scale anchoring}

\newcommand{\semmech}{calibrated semantic progress alignment}

\title{\raisebox{-0.22\height}{\includegraphics[height=1.22em]{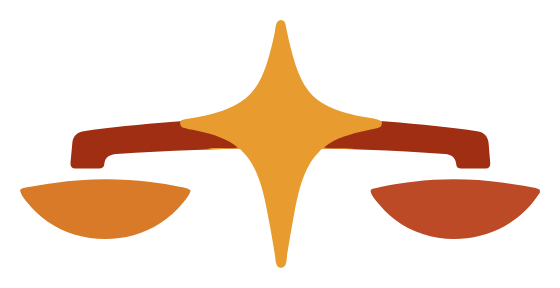}}\hspace{0.25em}Yuheng: Toward Universal Progress Modeling through In-Context Learning and Semantic Alignment}

\author[1,2,*]{Zihao Liu}
\author[1,3,*]{Chunfeng Wang}
\author[1,4]{Xinpan Meng}
\author[1,2]{Siyi Lang}
\author[1]{Yifan Bai}
\author[1]{Xinyuan Chang}
\author[1,5]{Zhiheng Ma}

\affiliation[1]{Primotion}
\affiliation[2]{Northwestern Polytechnical University}
\affiliation[3]{RWTH Aachen University}
\affiliation[4]{Institute of Automation, Chinese Academy of Sciences}
\affiliation[5]{Shenzhen University of Advanced Technology}
\contribution[*]{Equal contribution.}

\abstract{
Generalizable task progress estimation is essential for embodied policy refinement, execution monitoring, and robotics data curation. However, prevailing reward and progress models rely on elapsed time over successful demonstrations and unanchored language instructions, miscalibrating intermediate milestones and hallucinating rising progress on stalled or failed rollouts despite training on massive corpora. In this technical report, we introduce \method, a universal progress modeling framework built on \anchormech{} and \semmech{}. Given a task instruction and ordered keyframes with custom progress anchors from a single demonstration, \method localizes arbitrary query observations onto the reference scale through causal multimodal self-attention and a shared bilinear scale head without retraining on new tasks. Trained on a compact, rubric annotated corpus spanning successful, suboptimal, and failed executions, \method aligns states in semantic progress space rather than elapsed time, faithfully capturing stalls, retries, and regressions. On \bench, our out of distribution benchmark with a balanced outcome distribution, \methodtwentyseven sets a new open model state of the art, outperforming prior specialized reward models by a wide margin while requiring an order of magnitude fewer training trajectories, and surpassing proprietary GPT-6-Astra; our compact \methodfour likewise surpasses all prior open baselines, including Robometer-4B on the identical 4B backbone. Beyond strong benchmark calibration, \method unlocks steerable milestone scales, semantic alignment across trajectories, and zero-shot transfer across embodiments from human videos to grippers and dexterous hands for downstream robot learning.
\textcolor{\myemcolor}{Interactive demonstrations, code, and model weights are available on the \href{https://primotion-ai.github.io/yuheng.github.io/}{project website} and \href{https://github.com/primotion-ai/Yuheng-Inference}{GitHub repository}.}
}

\begin{document}
\maketitle

% Modular chapters stored in chapters/
\input{chapters/introduction}
\input{chapters/related_work}
\input{chapters/method}
\input{chapters/experiments}
\input{chapters/applications}

\section{Conclusion}\label{sec:conclusion}

Scaling robotic foundation models beyond imitation learning requires a reliable, generalizable measure of how much a manipulation trajectory has actually accomplished at every point in time. Prevailing reward and progress models approximate this signal through elapsed trajectory time or unanchored regression conditioned solely on language, a paradigm that conflates physical task advancement with clock time, leaves intermediate milestone scales ambiguous across users and embodiments, and relies on brute force scaling to hundreds of thousands or millions of trajectories while remaining brittle on unsuccessful executions. \method resolves these bottlenecks by reformulating progress estimation as in-context semantic alignment against explicit reference keyframes.

This reformulation rests on two complementary principles. First, \anchormech{} turns a single successful demonstration and its keyframe progress labels into a steerable visual ruler, allowing the adapted VLM backbone and shared bilinear scale head to score arbitrary query observations across tasks, milestone definitions, and embodiments without parameter updates. Second, \semmech{} replaces frame index supervision with rubric calibrated frame annotations and soft target interpolation in semantic progress space across trajectories, exposing the model directly to stalls, recoverable retries, and irreversible regressions during training. Together, these mechanisms allow \method to achieve leading progress calibration from only 16{,}321 curated trajectories, using up to 65 times fewer trajectories than million-scale generalist reward models.

Comprehensive evaluation on \bench across 90 unseen tasks, 1{,}101 trajectories with a balanced distribution of outcomes, and 58{,}272 pairwise comparisons demonstrates the effectiveness of this design. \methodtwentyseven attains 75.7\% average agreement and 9.8 pp MAE, surpassing the strongest prior specialized reward model by 10.8 percentage points and outperforming the frontier $>10\text{T}$-parameter proprietary model GPT-6-Astra in overall agreement, calibration on unsuccessful trajectories, and framewise error. At a compact 4B scale, \methodfour reaches 73.3\% average agreement and improves upon Robometer-4B by 9.8 percentage points on the identical vision--language backbone. Qualitative and capability analyses further confirm that \method adapts immediately to modified milestone scales, aligns matching semantic stages across executions of varying speed, and transfers a single human hand reference zero-shot to parallel jaw grippers and dexterous robotic hands.

Several limitations and extensions point to promising directions for future work. First, \method currently evaluates each query image independently conditioned on the reference keyframes without ingesting temporal motion windows on the query side; incorporating lightweight causal query context could disambiguate directional primitives such as cyclic stirring, shaking, or symmetric opening versus closing when static frames appear visually similar. Second, while single camera RGB conditioning transfers effectively across diverse kinematics, tight contact assembly tasks with severe visual occlusion may benefit from fusing wrist camera views, tactile signals, or proprioceptive force cues into the query representation. Finally, building on the universal progress calibration and cross-embodiment alignment established in this report, a compelling next step for future work is closing the loop in live robot learning---deploying \method across autonomous real world reinforcement learning fleets, standardized data collection systems, and large scale video curation pipelines toward continually improving embodied foundation models.

\bibliographystyle{unsrtnat}
\bibliography{sn-bibliography}

\clearpage
\input{chapters/appendix}

\end{document}

%% file: chapters/introduction.tex
\begin{figure}[!tp]
\centering
\includegraphics[width=\textwidth]{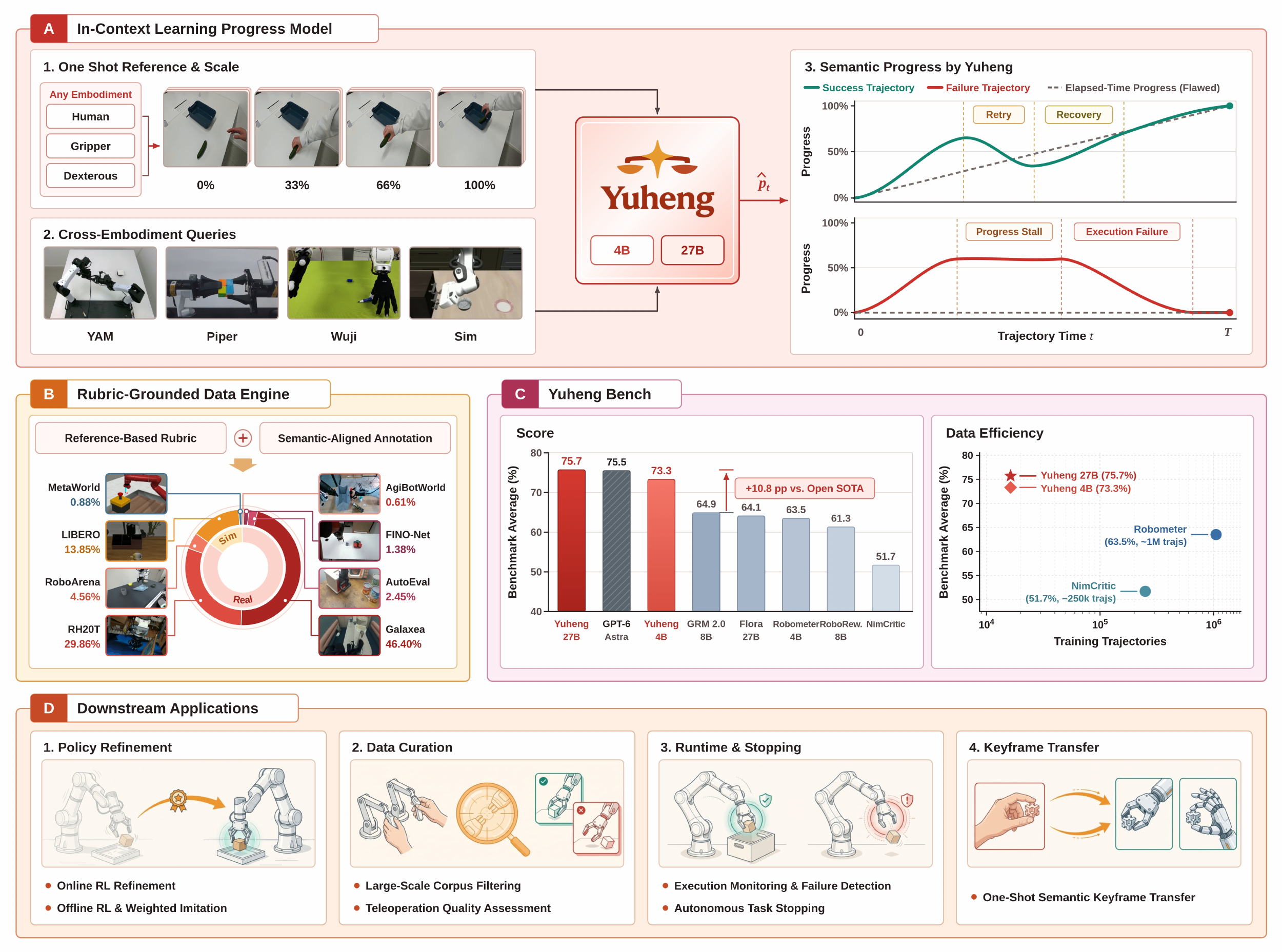}
\vspace{-6pt}
\caption{\textbf{Overview of \method: universal semantic progress modeling from a single reference demonstration.}
(\textbf{A})~\textbf{In-context progress model.} Given ordered keyframes and custom progress anchors from one demonstration of any embodiment, \method (\methodfour and \methodtwentyseven) scores cross-embodiment queries without task-specific retraining, capturing retries, recoveries, stalls, and failures rather than flawed elapsed time proxies.
(\textbf{B})~\textbf{Rubric grounded data engine.} Reference calibrated rubrics and semantic annotation curate a compact, verified corpus across eight real world and simulation datasets.
(\textbf{C})~\textbf{\bench{} evaluation.} Across unseen manipulation tasks with a balanced outcome distribution, \methodtwentyseven establishes a new state of the art and surpasses GPT-6-Astra, while \methodfour outperforms prior open baselines with far fewer training trajectories.
(\textbf{D})~\textbf{Downstream applications.} Calibrated semantic progress directly supports policy refinement (online RL and offline weighted imitation), data curation and teleoperation quality assessment, runtime monitoring and autonomous stopping, and one-shot semantic keyframe transfer (Sec.~\ref{sec:applications}).}
\label{fig:teaser}
\end{figure}

\section{Introduction}\label{sec:intro}

Advancing generalist robotic manipulation beyond behavior cloning requires models that do not merely imitate demonstrations, but understand how well a task is unfolding at every step~\cite{RT2,OpenXEmbodiment,pi0,pi05,intelligence2025pi06vla}. Across the modern robot learning lifecycle, dense task progress serves as a unifying supervisory currency: prior to training, it filters uncurated teleoperation and play corpora by separating purposeful advancement from erratic motion~\cite{budzianowski2025opengvl,memmel2025strap,leng2026flora}; during policy refinement, it supplies shaped process rewards for online reinforcement learning~\cite{wagenmaker2025steering,Luo2025HILSERL,li2025simplevlarl} and calibrated value or advantage weights for offline policy improvement~\cite{kostrikov2022offline,awr,intelligence2025pi06vla}; during deployment, it monitors live rollouts to detect stalls, trigger autonomous task stopping, and flag irreversible failures before timeouts or compounding errors occur~\cite{duan2024aha,gu2025safe,agia2025unpacking}; and across trajectories and embodiments, it enables one-shot semantic keyframe transfer (Fig.~\ref{fig:teaser}D). Realizing this vision across diverse open tasks, novel objects, and heterogeneous robot morphologies demands a universal progress model that generalizes out of the box without manual reward engineering or task-specific retraining.

Despite rapid progress in vision--language reward and value modeling~\cite{ma2023liv,ma2024generative,tan2025robodopamine,lee2026roboreward,robometer2026,leng2026flora,huang2026rynnvalue,fang2026densereward,xu2026nimcritic}, existing generalist progress estimators remain fundamentally limited when confronted with realistic robot executions of varying quality. These limitations stem from three structural mismatches in how task progress is formulated, supervised, and evaluated.
(1)~\textbf{Conflating elapsed trajectory time with task semantics corrupts progress under imperfect execution.} Prevailing approaches supervise dense progress using normalized frame index ($t/T$) or remaining duration along successful demonstrations, while collapsing failed episodes to flat zero credit or omitting them altogether~\cite{robometer2026,huang2026rynnvalue}. Such temporal proxies miss retries and recoveries on successful rollouts and erase intermediate partial advancement, progress stalls, and execution failure drops on failed attempts (Fig.~\ref{fig:teaser}A, right); as a result, models trained on them learn visual shortcuts tied to arm motion or video duration, predicting inflated, steadily rising progress at test time even when a robot hesitates before contact, stalls against an obstacle, or drops a grasped object.
(2)~\textbf{Unanchored language conditioning leaves task progress inherently ambiguous in scale.} Most existing reward foundation models predict progress conditioned solely on a text instruction and an observed trajectory~\cite{lee2026roboreward,robometer2026,leng2026flora}. Yet natural language specifies the terminal goal rather than how intermediate milestones partition credit along the way: in a drawer closing or assembly task, reaching and establishing contact may represent 20\% progress under one control rubric and 60\% under another. Without an explicit visual reference scale, such models face irreducible scale ambiguity across novel tasks and struggle to calibrate shared semantic stages across distinct embodiments such as human hands, parallel jaw grippers, and dexterous hands~\cite{chen2026host,Zakka2022XIRL}.
(3)~\textbf{Success bias in training and evaluation masks severe miscalibration on failed rollouts.} Prior efforts attempt to compensate for weak temporal labels through brute force dataset scaling, training on hundreds of thousands to over one million trajectories~\cite{robometer2026}. However, because demonstration repositories are overwhelmingly dominated by successful episodes~\cite{OpenXEmbodiment,khazatsky2024droid,tian2026position}, resulting models exhibit strong optimism bias: they assign inflated scores to incomplete trajectory prefixes and fail to distinguish recoverable hesitation from terminal failure, a weakness obscured by standard correlation metrics computed solely on successful rollouts~\cite{leng2026flora}.

To overcome these bottlenecks, we draw inspiration from the classical concept of measurement: quantifying a physical quantity requires comparing an unknown state against an explicit, calibrated standard. We introduce \method (named after \emph{Yuheng}, i.e., Alioth, the fifth star of the Big Dipper traditionally symbolizing balance and measurement), a universal progress modeling framework built on \anchormech{} and \semmech{} (Fig.~\ref{fig:teaser}A--B). Rather than treating progress estimation as unanchored temporal regression, \method reformulates it as an in-context semantic alignment problem conditioned on a single reference demonstration and a custom progress scale. This formulation is governed by two core principles.

First, \textbf{task progress should be anchored explicitly in context by reference keyframes} (\anchormech{}). Given a language instruction and an ordered sequence of $K$ keyframes $(k_1, \dots, k_K)$ selected from one successful demonstration together with monotonic progress anchors $(p_1, \dots, p_K)$, \method jointly encodes the instruction, reference keyframes, explicit scale prompts, and current query frame $x_t$ within a causal multimodal vision--language backbone. A shared bilinear scale projection head computes elementwise anchor--query interactions conditioned on each anchor's scalar value $p_i$, outputs a categorical distribution $\bm{\pi} = (\pi_1, \dots, \pi_K)$ over the $K$ reference milestones, and decodes continuous relative progress via expected value $\hat{p}_t = \sum_{i=1}^K \pi_i p_i$. By converting a single demonstration into an explicit visual ruler, \method resolves milestone ambiguity at test time, adapts instantaneously to arbitrary custom progress scales without retraining, and transfers across embodiments without target demonstrations.

Second, \textbf{supervision must align states in semantic progress space rather than trajectory time} (\semmech{}). To sever the spurious link between elapsed time and task accomplishment, we develop a two-stage rubric annotation pipeline powered by frontier multimodal reasoning models. This pipeline first establishes a frozen task specific semantic rubric and unified reference keyframes; it then annotates dense semantic progress and execution outcomes across successful, suboptimal, and failed trajectories with automated consistency verification, avoiding elapsed time interpolation, monotonicity forcing, or endpoint clamping. During training, \method pairs deduplicated reference anchors from successful or suboptimal demonstrations with query frames drawn across trajectories of all execution outcomes, assigning piecewise linear soft classification targets strictly according to semantic progress distance rather than frame index. Coupled with multimodal attention adaptation across two complementary scales---\methodfour (built on Qwen3-VL-4B~\cite{qwen3vl}) and parameter efficient LoRA adaptation in \methodtwentyseven (built on Qwen3.8-27B~\cite{qwen38})---this supervision eliminates optimism bias with high data efficiency from a compact curated corpus spanning diverse simulation and physical robot datasets.

To evaluate detailed progress modeling without success bias, we construct \bench, a contamination free out of distribution (OOD) benchmark spanning unseen manipulation tasks with a balanced distribution of successful, suboptimal, failed, and truncated trajectories, assessed via within-trajectory agreement, cross-trajectory agreement, terminal outcome agreement, and framewise error. Across frontier proprietary VLMs and specialized open reward models (Fig.~\ref{fig:teaser}C), \methodtwentyseven establishes a new state of the art, outperforming prior specialized baselines by over 10 percentage points while halving frame error and exceeding proprietary GPT-6-Astra in overall agreement and calibration. At a compact 4B scale, \methodfour likewise surpasses all prior open baselines---including Robometer-4B on the identical vision--language backbone---while requiring far fewer training trajectories.

In summary, this technical report makes three main contributions:
(1)~\textbf{In-context semantic progress formulation and architecture.} We introduce \method (\methodfour and \methodtwentyseven), which pairs causal multimodal reference conditioning with a variable-cardinality bilinear scale projection head to unify rapid task generalization from a single demonstration, steerable custom progress scales, semantic alignment across trajectories, and direct transfer across embodiments.
(2)~\textbf{Data-efficient rubric curation and semantic-space training.} We design a two-stage rubric annotation pipeline and a cross-trajectory soft-target sampling scheme that decouple progress learning from elapsed time, substantially improving cross-trajectory alignment and terminal outcome calibration while halving frame error over \emph{Yuheng w/o Semantic Labels} (trained with same-trajectory $t/T$ interpolation on the identical architecture) from a compact curated corpus.
(3)~\textbf{Balanced OOD benchmark and comprehensive evaluation.} We construct \bench with a balanced trajectory distribution spanning successful, suboptimal, failed, and truncated executions across unseen manipulation tasks, and demonstrate that both \methodtwentyseven and \methodfour establish a new state of the art among open models, with \methodtwentyseven further surpassing proprietary GPT-6-Astra in overall agreement and both models achieving tighter framewise calibration.

%% file: chapters/related_work.tex
\section{Related Work}\label{sec:related_work}

\subsection{Reward and Progress Modeling for Robotic Manipulation}\label{sec:rw_reward}

Learning visual reward and value functions directly from videos has long been pursued to eliminate manual reward engineering in physical robot environments~\cite{nair2022r3m,ma2023eureka,rocamonde2024visionlanguage}. Early representation methods learn time-contrastive or goal-conditioned visual embeddings, video--language similarities, or temporal ranking functions as proxy rewards, including TCN~\cite{Sermanet2018TCN}, XIRL~\cite{Zakka2022XIRL}, GraphIRL~\cite{GraphIRL}, VIP~\cite{ma2023vipuniversalvisualreward}, LIV~\cite{ma2023liv}, RoboCLIP~\cite{roboclip}, and Rank2Reward~\cite{yang2024rank}. While effective within narrow domains, embedding distances often lack calibrated $[0,1]$ scale semantics and degrade when viewpoints, lighting, or robot morphologies shift across tasks, motivating language- and subtask-aware reward models~\cite{alakuijala2024videolanguagecritictransferablereward,kim2025reds,hung2025victor}. Preference reinforcement learning methods such as T-REX~\cite{brown2019trex}, PEBBLE~\cite{lee2021pebble}, and RL-VLM-F~\cite{wangrlvlmf2024} infer rewards from pairwise trajectory comparisons~\cite{christiano2017rlhf,Bradley1952RankAO}, but typically train separate critics per task rather than a single generalist progress estimator.

Recent efforts scale reward and progress modeling onto pretrained vision--language backbones across large diverse robot corpora~\cite{zhang2025rewind,zhai2025VLAC,tan2025robodopamine,lee2026roboreward,robometer2026,leng2026flora,huang2026rynnvalue}. RoboReward~\cite{lee2026roboreward} trains vision--language outcome critics via counterfactual relabeling and temporal clipping to score discrete episode completion. RynnValue~\cite{huang2026rynnvalue} scales value foundation models by predicting temporal distance (remaining time to completion), whereas SARM~\cite{chen2025sarm} and ProgressLM~\cite{zhang2026progresslmprogressreasoningvisionlanguage} incorporate stage reasoning. DenseReward~\cite{fang2026densereward} synthesizes manipulation failures in simulation to learn dense frame-level rewards, while NimCritic~\cite{xu2026nimcritic} performs action-aware quantification for predictive long-horizon progress modeling. Closer to our setting, Robometer~\cite{robometer2026} combines uniform local temporal progress supervision ($t/T$) with trajectory preference comparisons over one million trajectories (RBM-1M), and FLORA~\cite{leng2026flora} unifies outcome, retrospective dense step quality, and rubric conditioned reward prediction in a 27B VLM data engine. Despite their scale, unanchored models that map instructions and video prefixes directly to progress remain vulnerable to milestone ambiguity on unseen tasks, while temporal $t/T$ supervision on expert demonstrations induces optimism bias on stalled or regressing trajectories. \method addresses both limitations simultaneously by anchoring progress onto explicit in-context reference keyframes and supervising alignment with calibrated semantic progress labels rather than elapsed time.

\subsection{In-Context Visual Conditioning and Embodiment Alignment}\label{sec:rw_icl}

Prompting foundation models with visual examples at inference time offers a powerful alternative to fine-tuning for every new task~\cite{ma2024generative,tan2025robodopamine,chen2026host}. In value estimation, Generative Value Learning (GVL)~\cite{ma2024generative} and OpenGVL~\cite{budzianowski2025opengvl} cast progress prediction as zero-shot or in-context frame ordering by prompting general-purpose VLMs to regress shuffled frame percentages, while TOPReward~\cite{chen2026topreward} extracts token probabilities from pretrained VLMs as rewards without training. Robo-Dopamine~\cite{tan2025robodopamine} and VLAC~\cite{zhai2025VLAC} condition process reward models on reference images or frame pairs to predict relative progress hops for precision manipulation. However, autoregressive text generation of numeric percentages suffers from discretization noise and poor calibration on open models, while iterative hop accumulation can drift over long horizons when rollouts stall or reverse~\cite{robometer2026}. In contrast, \method performs causal multimodal self-attention in a single pass over an interleaved sequence of $K$ labeled reference keyframes and the query frame, decoding continuous relative progress through a dedicated bilinear scale projection head trained with soft classification targets in progress space~\cite{bellemare2017distributional,farebrother2024stop}.

Our work also connects to correspondence across embodiments and observational learning from human demonstrations~\cite{dwibedi2019tcc,hadji2021,Vid2Robot,EgoScale,MimicPlay,chen2026host}. Temporal alignment methods such as TCC~\cite{dwibedi2019tcc}, Soft-DTW~\cite{SoftDTW,hadji2021}, XIRL~\cite{Zakka2022XIRL}, and HOST~\cite{chen2026host} map human videos and robot executions onto shared task progress manifolds to transfer manipulation procedures across morphologies. Whereas policy centric frameworks couple aligned progress directly to embodiment specific action generation, \method isolates universal semantic progress measurement itself: given a single human or robot demonstration with custom milestone values, \method transfers a shared, steerable progress scale directly across parallel jaw grippers, dexterous hands, and human manipulation.

\subsection{Embodied Policy Refinement, Execution Monitoring, and Data Curation}\label{sec:rw_downstream}

The practical value of a generalist progress model lies in its downstream consumers across policy improvement, safety monitoring, and dataset curation~\cite{intelligence2025pi06vla,robometer2026,leng2026flora}. For online policy improvement, reinforcement learning atop pretrained visuomotor, vision--language--action (VLA), and world action model (WAM) policies~\cite{DiffusionPolicy,OpenVLA,pi0,pi05,WallOSS,ye2026dreamzero}---such as latent-space Diffusion Steering RL (DSRL)~\cite{wagenmaker2025steering}, VLA-RL~\cite{lu2025vlarlmasterfulgeneralrobotic}, and SimpleVLA-RL~\cite{li2025simplevlarl}---as well as real-world human-in-the-loop RL~\cite{Luo2025HILSERL}, requires dense, calibrated progress rewards and reliable success indicators to avoid manual reward engineering and success labeling. For offline policy refinement on mixed quality experience, Implicit Q-Learning (IQL)~\cite{kostrikov2022offline}, Advantage Weighted Regression (AWR / AW-BC)~\cite{awr}, value-guided policy steering~\cite{nakamoto2024steering}, and experience driven VLA refinement such as $\pi^*_{0.6}$ (RECAP)~\cite{intelligence2025pi06vla,ghasemipour2025selfimproving} rely on dense value estimates and quality weights to extract strong behaviors without diverging toward suboptimal trajectories. Beyond policy optimization, runtime failure monitors~\cite{du2023vision,duan2024aha,gu2025safe,lu2025robofac,agia2025unpacking,lin2025failsafe,dai2025racer} and test time verifiers~\cite{kwok2025robomonkey,kwok2026llmverifier} track execution progress to flag stalls or regressions, while data curation pipelines~\cite{memmel2025strap,zhang2026scizor,agia2025cupid,budzianowski2025opengvl} filter noisy robotic datasets collected across institutions~\cite{OpenXEmbodiment,khazatsky2024droid,jiang2025galaxea,fang2023rh20t,atreya2025roboarena,contributors2024agibotworldrepo}. \method provides a single, unified semantic progress backbone engineered to support these downstream paradigms (Sec.~\ref{sec:applications}).

%% file: chapters/method.tex
\section{Method}\label{sec:method}

\begin{figure}[!tp]
\centering
\includegraphics[width=\textwidth]{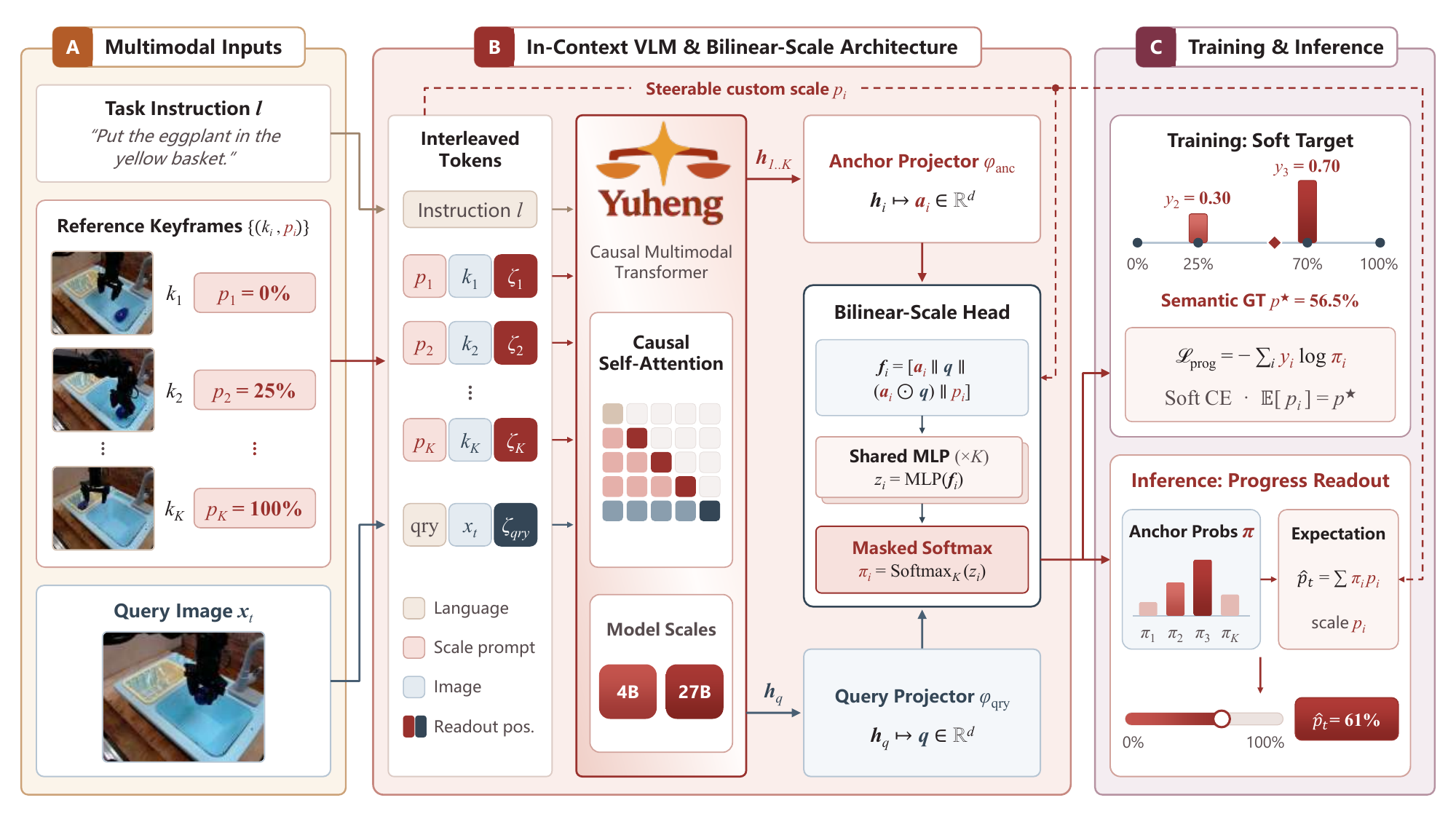}
\vspace{-6pt}
\caption{\textbf{Overview of the \method in-context progress modeling framework.}
(\textbf{A})~\method takes a natural language task instruction $l$, $K$ ordered keyframes $\{(k_i, p_i)\}_{i=1}^{K}$ paired with monotonic progress labels, and a current query image $x_t$.
(\textbf{B})~Inputs are formatted into an interleaved multimodal token sequence with readout positions $(\zeta_1, \dots, \zeta_K, \zeta_{\mathrm{qry}})$ and encoded jointly by the \method VLM (4B or 27B) to extract anchor hidden states $(\bm{h}_1, \dots, \bm{h}_K)$ and a terminal query hidden state $\bm{h}_q$; dedicated projectors ($\phi_{\mathrm{anc}}, \phi_{\mathrm{qry}}$) map them to $(\bm{a}_i, \bm{q})$, which a shared bilinear scale head scores over $[\bm{a}_i \,\|\, \bm{q} \,\|\, (\bm{a}_i \odot \bm{q}) \,\|\, p_i]$ to produce categorical keyframe probabilities $\bm{\pi}$.
(\textbf{C})~During training (top), $\bm{\pi}$ is optimized via soft cross-entropy against piecewise linear targets $\bm{y}$ interpolated in reference progress space around $p^{\star}$ (preserving $\mathbb{E}_{i \sim \bm{y}}[p_i] = p^{\star}$); at inference time (bottom), continuous relative progress is decoded as the expected value $\hat{p}_t = \sum_{i=1}^{K} \pi_i p_i$, with steerable scale $p_i$ simultaneously conditioning prompts, head interactions, and readout.}
\label{fig:framework}
\end{figure}

\subsection{Problem Formulation and Framework Overview}\label{sec:prob_formulation}

\textbf{Problem formulation.}
Consider a robotic manipulation task specified by a natural language instruction $l$. Let $\tau^{\mathrm{ref}} = (o^{\mathrm{ref}}_1, \dots, o^{\mathrm{ref}}_{T_{\mathrm{ref}}})$ denote a reference trajectory demonstrating monotonic task advancement, from which an operator or automated selector extracts $K$ chronologically ordered visual keyframes $\mathcal{K} = (k_1, \dots, k_K)$ ($K \ge 2$) together with custom monotonic progress anchors $\mathcal{P} = (p_1, \dots, p_K)$ satisfying
{\small
\begin{equation}
0 = p_1 < p_2 < \dots < p_K = 1.
\label{eq:anchor_scale}
\end{equation}
}
Given an arbitrary query trajectory $\tau = (x_1, \dots, x_T)$ executing the same task $l$ under potentially different execution speed, object pose, terminal outcome (successful, suboptimal, or failed), or physical embodiment (parallel jaw gripper, human hand, or dexterous hand), our objective is to learn a universal progress estimator $F_\theta$ that maps each query frame $x_t$ to its continuous relative semantic progress $\hat{p}_t \in [0, 1]$ on the scale defined by $(\mathcal{K}, \mathcal{P})$:
{\small
\begin{equation}
\hat{p}_t = F_\theta\big(x_t \;\big|\; l,\, \{(k_i, p_i)\}_{i=1}^{K}\big), \qquad t \in \{1, \dots, T\}.
\label{eq:problem_def}
\end{equation}
}
Crucially, $F_\theta$ operates in an online, frame independent query regime without future lookahead, requiring no retraining for individual tasks, post hoc temporal smoothing, or artificial endpoint clamping ($\hat{p}_1 \equiv 0$ or $\hat{p}_T \equiv 1$). Consequently, when $\tau$ stalls at an intermediate milestone or undoes an earlier manipulation step, $\hat{p}_t$ must faithfully plateau or decrease rather than drifting upward with elapsed time.

\textbf{Framework overview.}
As depicted in Fig.~\ref{fig:framework}, \method realizes Eq.~\eqref{eq:problem_def} through three synergistic design pillars organized from forward architecture to training and data curation:
(1)~an \emph{in-context multimodal VLM and variable cardinality bilinear scale projection head} (Sec.~\ref{sec:arch}, Fig.~\ref{fig:framework}A--B and C, bottom) that jointly attends across the task instruction, scaled reference keyframes, and current query frame to decode expected progress on arbitrary anchor schedules;
(2)~a \emph{diverse outcome sampling and semantic space soft target training protocol} (Sec.~\ref{sec:training}, Fig.~\ref{fig:framework}C, top) that aligns query observations from successful, suboptimal, and failed executions directly in normalized semantic progress space; and
(3)~a \emph{two-stage rubric semantic annotation pipeline} (Sec.~\ref{sec:data_engine}, Fig.~\ref{fig:data_engine}) that curates 16{,}321 verified trajectories across eight manipulation corpora with dense, nonmonotonic ground truth free of elapsed time leakage.

\subsection{In-Context Multimodal Architecture and Bilinear Scale Head}\label{sec:arch}

\textbf{Structured interleaved multimodal tokenization.}
Unlike unanchored video reward models that ingest only a language prompt and target frames, or naive prompts that list keyframes and numeric labels separately, \method interleaves the task instruction $l$, each paired reference milestone $(p_i, k_i)$, and the query observation $x_t$ into a single causal token stream (Fig.~\ref{fig:framework}A--B). Specifically, each reference keyframe $k_i$ is preceded immediately by its ordinal index and declared percentage progress $100 \cdot p_i$ and enclosed by dedicated visual boundary tokens within an explicit reference sequence header, unambiguously separating the reference scale from the appended query image $x_t$ and terminal readout prompt (see Appendix~\ref{sec:app_prompt} for the complete prompt template). Under causal self-attention, the representation extracted at the $i$-th closing image boundary token (index $\zeta_i$) contextualizes keyframe $k_i$ with the task goal $l$, its declared progress $p_i$, and all preceding milestones $1, \dots, i-1$, while the final valid token (index $\zeta_{\mathrm{qry}}$) attends jointly across the entire reference ruler and the query image $x_t$.

\textbf{Vision--language backbones and parameter efficient adaptation.}
We instantiate the \method VLM backbone at two complementary model scales (Fig.~\ref{fig:framework}B) to serve both fast closed loop control and maximum precision offline evaluation:
(1)~\methodfour builds on Qwen3-VL-4B~\cite{qwen3vl} (hidden dimension $D = 2{,}560$, 36 full self-attention blocks), freezing the pretrained vision transformer while fine-tuning all language model self-attention and feedforward layers together with the progress head.
(2)~\methodtwentyseven scales to Qwen3.8-27B~\cite{qwen38} ($D = 5{,}120$), a 64 layer hybrid multimodal architecture alternating three gated linear attention blocks with one full causal self-attention block (16 full attention layers total). To adapt fine visual correspondence across frames while preserving pretrained world knowledge, we freeze the entire 27B backbone (vision encoder, embeddings, MLPs, and all 48 linear attention layers) and attach rank 256 Low Rank Adaptation (LoRA)~\cite{LoRA} modules ($\alpha = 64$, scaling $\alpha/r = 0.25$, dropout $0.05$) exclusively to the query, key, value, and output attention projections ($W_Q, W_K, W_V, W_O$) of the 16 full attention layers, updating only 167.8M LoRA parameters plus 2.8M head parameters (approximately $0.65\%$ of total weights).

\textbf{Variable cardinality bilinear scale projection head.}
Autoregressive text generation of numeric percentages suffers from token discretization artifacts, whereas fixed bin classification heads cannot adapt when users vary the number of keyframes $K$ or assign irregular milestone scales. We therefore introduce a shared bilinear scale projection head operating directly on the final layer hidden states $H \in \mathbb{R}^{L \times D}$. Let $\bm{h}_i = H[\zeta_i] \in \mathbb{R}^D$ ($i = 1, \dots, K$) and $\bm{h}_q = H[\zeta_{\mathrm{qry}}] \in \mathbb{R}^D$ denote the extracted anchor and query hidden states. Dedicated LayerNorm, Linear, and GELU projectors map both into a $d$ dimensional alignment subspace ($d = 256$):
{\small
\begin{equation}
\bm{a}_i = \phi_{\mathrm{anc}}(\bm{h}_i) = \mathrm{GELU}\big(W_a\,\mathrm{LN}(\bm{h}_i) + \bm{b}_a\big) \in \mathbb{R}^d, \qquad
\bm{q} = \phi_{\mathrm{qry}}(\bm{h}_q) = \mathrm{GELU}\big(W_q\,\mathrm{LN}(\bm{h}_q) + \bm{b}_q\big) \in \mathbb{R}^d.
\label{eq:proj}
\end{equation}
}
For each anchor $i \in \{1, \dots, K\}$, we concatenate the projected anchor and query representations, their multiplicative Hadamard alignment $\bm{a}_i \odot \bm{q}$, and the explicit scalar milestone position $p_i \in [0, 1]$ into a $(3d + 1)$ dimensional interaction vector $\bm{f}_i \in \mathbb{R}^{769}$, which is scored by a shared two layer MLP:
{\small
\begin{equation}
\bm{f}_i = \big[\, \bm{a}_i \;\big\|\; \bm{q} \;\big\|\; (\bm{a}_i \odot \bm{q}) \;\big\|\; p_i \,\big] \in \mathbb{R}^{3d + 1}, \qquad
z_i = \bm{w}_2^{\top}\,\mathrm{GELU}(W_1 \bm{f}_i + \bm{b}_1) + b_2 \in \mathbb{R}.
\label{eq:scorer}
\end{equation}
}
Because $(W_1, \bm{b}_1, \bm{w}_2, b_2)$ are shared across all $i$, the head handles arbitrary anchor cardinalities $K \in \{2, \dots, K_{\max}\}$ seamlessly. Masked softmax normalization over active anchors (masking padded minibatch slots $i > K$ to $-\infty$) yields calibrated keyframe classification probabilities:
{\small
\begin{equation}
\pi_i = \frac{\exp(z_i)}{\sum_{j=1}^{K} \exp(z_j)}, \qquad i = 1, \dots, K.
\label{eq:softmax}
\end{equation}
}

\textbf{Scale adaptive inference readout.}
At inference time (Fig.~\ref{fig:framework}C, bottom), given any unseen task instruction $l$, $K$ reference keyframes $(k_1, \dots, k_K)$ from one demonstration, and any chosen monotonic scale $0 = p_1 < p_2 < \dots < p_K = 1$, \method predicts continuous relative progress via expected value over the reference scale:
{\small
\begin{equation}
\hat{p}_t = \sum_{i=1}^{K} \pi_i \, p_i \in [0, 1].
\label{eq:inference}
\end{equation}
}
When the reference keyframes span a partial task subsegment $[p_{\mathrm{start}}, p_{\mathrm{end}}] \subset [0, 1]$, user-supplied anchor values are affinely normalized onto $[0, 1]$ prior to conditioning and linearly mapped back to $[p_{\mathrm{start}}, p_{\mathrm{end}}]$ after Eq.~\eqref{eq:inference}.
Because $p_i$ simultaneously conditions the multimodal prompt tokens, the bilinear scale interaction feature $\bm{f}_i$ in Eq.~\eqref{eq:scorer}, and the expected value aggregation in Eq.~\eqref{eq:inference}, modifying $(p_1, \dots, p_K)$ at test time immediately reshapes the predicted progress profile to follow the user's desired task rubric without updating a single parameter.

\subsection{Semantic Space Sampling and Soft Target Training}\label{sec:training}

\textbf{Dynamic reference sampling and plateau deduplication.}
To train \method to generalize across diverse keyframe counts and irregular anchor spacings, each training sample selects one of the eight data sources uniformly so smaller manipulation domains are not overwhelmed by large repositories, and draws a reference trajectory $\tau^{\mathrm{ref}}$ with a verified successful or suboptimal outcome. An initial count $K_0 \sim \mathcal{U}\{3, \dots, 11\}$ of chronological frames $t_1 < t_2 < \dots < t_{K_0}$ with raw semantic labels $(\tilde{p}_{t_1}, \dots, \tilde{p}_{t_{K_0}})$ is drawn and passed through two structural filters:
(1)~\emph{Monotonicity verification}, which rejects any draw with $\tilde{p}_{t_m} > \tilde{p}_{t_{m+1}}$ so reference rulers always progress forward; and
(2)~\emph{Plateau deduplication}, which collapses consecutive stationary pauses satisfying $\tilde{p}_{t_{m+1}} - \tilde{p}_{t_m} < \delta_{\mathrm{plat}}$ ($\delta_{\mathrm{plat}} = 0.01$) by retaining a single uniformly sampled representative frame per plateau. Draws with fewer than two surviving keyframes ($K < 2$) or zero span ($\tilde{p}_{t_K} - \tilde{p}_{t_1} \le 0$) are discarded, with trajectory feasibility screened in advance via $\mathcal{O}(T^2)$ dynamic programming over trajectory metadata prior to image decoding.

\textbf{Cross trajectory query pairing across outcomes.}
Let $[\tilde{a}, \tilde{b}] = [\tilde{p}_{t_1}, \tilde{p}_{t_K}]$ ($\tilde{a} < \tilde{b}$) denote the raw semantic progress span of the surviving reference anchors. Rather than sampling queries only from $\tau^{\mathrm{ref}}$, we draw a query trajectory $\tau^{\mathrm{qry}}$ uniformly across \emph{all} trajectories of the same task in the training split, explicitly spanning successful, suboptimal, and failed rollouts, and sample a frame $x_t$ whose raw semantic progress $\tilde{p}^{\star}$ lies in $[\tilde{a}, \tilde{b}]$. Both anchor values and query target are affinely normalized onto the local reference segment:
{\small
\begin{equation}
p_i = \frac{\tilde{p}_{t_i} - \tilde{a}}{\tilde{b} - \tilde{a}} \in [0, 1] \quad (i = 1, \dots, K), \qquad
p^{\star} = \frac{\tilde{p}^{\star} - \tilde{a}}{\tilde{b} - \tilde{a}} \in [0, 1].
\label{eq:segment_norm}
\end{equation}
}
This local segment normalization naturally synthesizes rich, irregular anchor scales $(p_1, \dots, p_K)$ during training while teaching \method to evaluate partial task subsegments as well as full episodes.

\textbf{Piecewise linear soft classification objective.}
Direct scalar regression ($L_1$/$L_2$) onto $p^{\star}$ suffers from representation collapse and poor calibration on high capacity VLMs~\cite{bellemare2017distributional,farebrother2024stop}. Instead, we supervise the categorical distribution $\bm{\pi}$ with a soft target $\bm{y} = (y_1, \dots, y_K) \in \Delta^{K-1}$ spanning two adjacent anchors, interpolated strictly in normalized semantic progress space (Fig.~\ref{fig:framework}C, top). Because plateau deduplication guarantees strictly increasing training anchors ($0 = p_1 < p_2 < \dots < p_K = 1$), for $p^{\star} \in [0, 1]$ bracketed by adjacent anchors $p_\ell \le p^{\star} \le p_{\ell+1}$ ($p_\ell < p_{\ell+1}$), we compute the interpolation ratio $\alpha = (p^{\star} - p_\ell)/(p_{\ell+1} - p_\ell) \in [0, 1]$ and assign target mass
{\small
\begin{equation}
y_i =
\begin{cases}
1 - \alpha, & \text{if } i = \ell, \\
\alpha, & \text{if } i = \ell + 1, \\
0, & \text{otherwise},
\end{cases}
\label{eq:soft_target}
\end{equation}
}
reducing to a point mass $y_\ell = 1$ when $p^{\star} = p_\ell$. Crucially, this soft target preserves exact expectation consistency with the inference readout in Eq.~\eqref{eq:inference}:
{\small
\begin{equation}
\mathbb{E}_{i \sim \bm{y}}[p_i] = (1 - \alpha)\,p_\ell + \alpha\,p_{\ell+1} = p^{\star},
\label{eq:exp_consistency}
\end{equation}
}
ensuring that minimizing soft cross-entropy drives the expected-value prediction $\hat{p}_t = \sum_{i=1}^{K} \pi_i p_i$ to converge unbiasedly to $p^{\star}$. For instance, under normalized anchors $[0\%, 25\%, 70\%, 100\%]$ and query semantic progress $p^{\star} = 56.5\%$, interpolating on $[25\%, 70\%]$ gives $\alpha = (0.565 - 0.25)/(0.70 - 0.25) = 0.70$, yielding $\bm{y} = [0,\, 0.30,\, 0.70,\, 0]$ rather than uniform index weights. All trainable parameters are optimized jointly via soft cross-entropy:
{\small
\begin{equation}
\mathcal{L}_{\mathrm{prog}}(\theta) = -\sum_{i=1}^{K} y_i \log \pi_i.
\label{eq:loss}
\end{equation}
}

\subsection{Rubric Semantic Annotation and Data Curation}\label{sec:data_engine}

\begin{figure}[!tp]
\centering
\includegraphics[width=\textwidth]{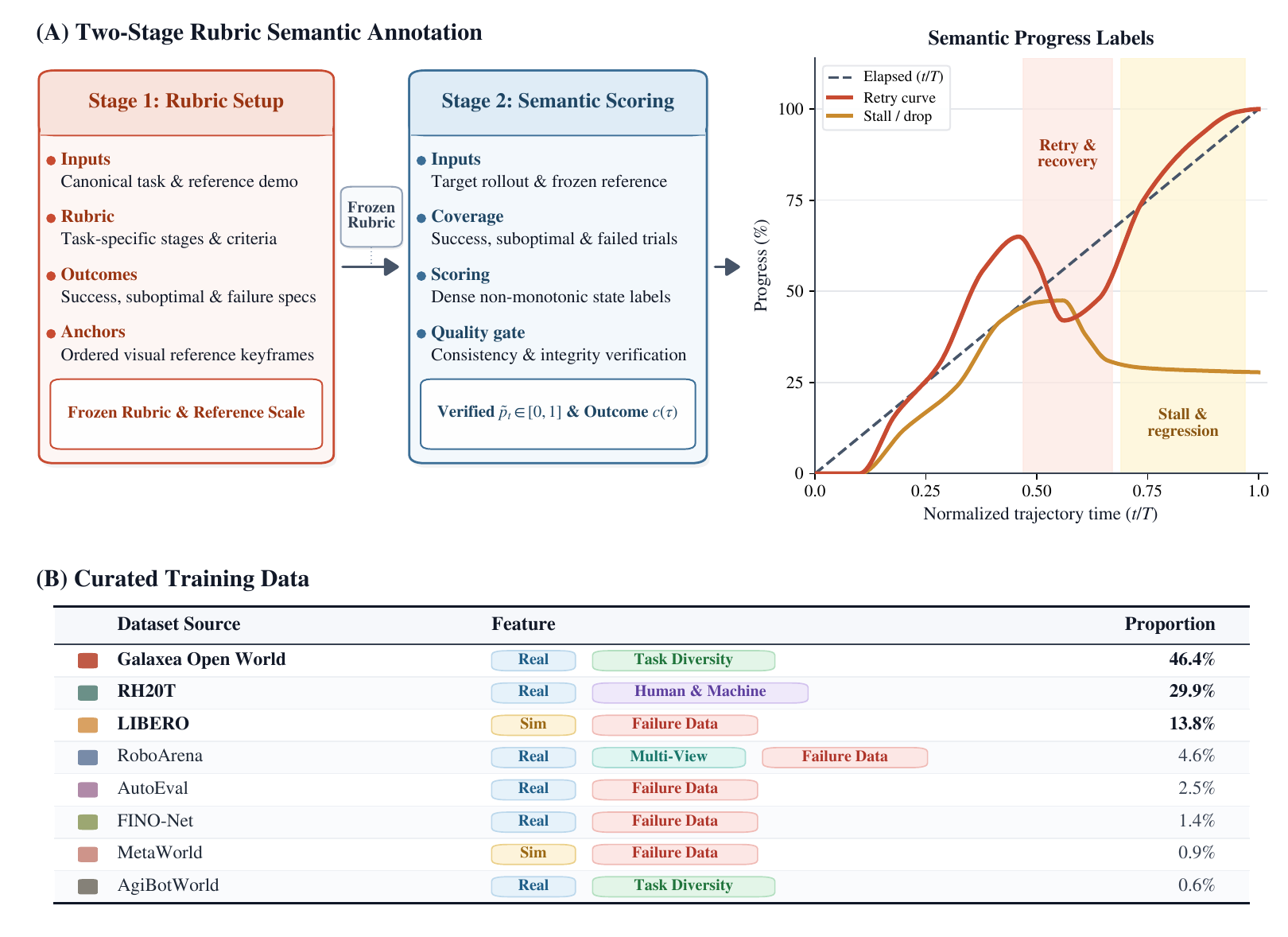}
\vspace{-6pt}
\caption{\textbf{Two-stage rubric semantic curation pipeline and diverse training corpus.}
(\textbf{A})~Our annotation pipeline decouples semantic progress calibration from elapsed time ($t/T$) in two stages: Stage~1 establishes a frozen task-specific semantic rubric and reference keyframes, while Stage~2 scores dense per-frame task accomplishment across successful, suboptimal, and failed rollouts with automated consistency verification and without time interpolation, monotonicity forcing, or endpoint clamping.
(\textbf{B})~Distribution of the 16{,}321 verified trajectories curated across eight real robot and simulation manipulation datasets.}
\label{fig:data_engine}
\end{figure}

Supervising Eq.~\eqref{eq:soft_target} without temporal bias requires dense trajectory annotations grounded in physical state transitions rather than frame indices. Manual frame scoring does not scale across diverse domains, whereas unconstrained single pass VLM prompting yields inconsistent milestone boundaries across rollouts. We resolve this bottleneck through a two-stage rubric annotation pipeline (Fig.~\ref{fig:data_engine}A) powered by frontier multimodal reasoning models that separates task-level scale calibration from per-trajectory semantic scoring.

\textbf{Stage 1: Task rubric calibration and reference setup.}
For each manipulation task, Stage~1 constructs and freezes a structured semantic rubric defining the required manipulation stages, verifiable completion criteria, and outcome distinctions (successful, suboptimal, and failed), together with a compact set of strictly progressing visual reference keyframes spanning the task from initial state to completion. Anchoring every task to a frozen rubric and shared reference keyframes ensures consistent progress semantics across diverse rollouts while preventing irrelevant arm motion or elapsed time from accruing spurious credit.

\textbf{Stage 2: Dense semantic scoring and quality verification.}
Conditioned on the frozen Stage~1 rubric and reference keyframes, Stage~2 evaluates each target trajectory across its stored frames to produce dense, calibrated semantic progress labels $\tilde{p}_t \in [0, 1]$ together with a verified semantic terminal outcome $c(\tau) \in \{\text{success}, \text{suboptimal}, \text{failure}\}$. Automated consistency and quality checks validate every annotated trajectory without applying elapsed-time interpolation, monotonicity forcing, temporal smoothing, or endpoint clamping. Consequently, stalled grasps naturally plateau, recoverable slips dip and rebound, and irreversible drops fall sharply toward zero.

\textbf{Benchmark disjoint curation and split hygiene.}
Rather than training on noisy million trajectory dumps, we curate high precision semantic trajectories from the processed diverse robot pools in Robometer~\cite{robometer2026}, strictly excluding all OOD evaluation benchmarks as well as external reward model training splits (e.g., RoboReward) to guarantee zero contamination. Automated quality checks reject incomplete, ambiguous, or corrupted trajectory records. As shown in Fig.~\ref{fig:data_engine}B, this yields \textbf{16{,}321} verified trajectories across eight physical robot and simulation sources: Galaxea Open World~\cite{jiang2025galaxea} (7{,}574, 46.4\%), RH20T~\cite{fang2023rh20t} (4{,}873, 29.9\%), LIBERO~\cite{liu2023libero} (2{,}260, 13.8\%), RoboArena~\cite{atreya2025roboarena} (745, 4.6\%), AutoEval~\cite{zhou2025autoeval} (400, 2.5\%), FINO-Net~\cite{inceoglu2021finonet} (225, 1.4\%), MetaWorld~\cite{yu2020metaworld} (144, 0.9\%), and AgiBotWorld~\cite{contributors2024agibotworldrepo} (100, 0.6\%). A deterministic validation split holds out $1\%$ of trajectories per task, shared identically across \methodfour and \methodtwentyseven so validation trajectories never appear as either references or queries during training.

%% file: chapters/experiments.tex
\section{Experiments}\label{sec:experiments}

We evaluate \method through a structured progression of quantitative benchmarks, qualitative case studies, capability analyses, and controlled mechanistic ablations. Specifically, our experiments are organized to answer four core questions:
(1)~Does \method achieve superior dense progress accuracy, calibration across trajectories, and failure sensitivity on unseen OOD tasks compared with frontier proprietary VLMs and specialized million trajectory reward models (Sec.~\ref{sec:main_results} and Sec.~\ref{sec:qualitative})?
(2)~Can \method dynamically adapt to custom milestone scales, align matching semantic stages across variable speed trajectories, and transfer a shared progress ruler directly across embodiments and manipulation domains from a single demonstration (Sec.~\ref{sec:capabilities})?
(3)~How much do calibrated semantic supervision and multimodal attention adaptation individually contribute over elapsed time ($t/T$) supervision and frozen backbone probing (Sec.~\ref{sec:ablations})?
(4)~How does \method scale across 4B and 27B backbones and training schedules while operating at up to 65 times greater data efficiency than prior generalist reward models (Sec.~\ref{sec:main_results} and Sec.~\ref{sec:ablations})?

\subsection{Experimental Setup, \bench, and Baselines}\label{sec:exp_setup}

\textbf{\bench construction.}
Existing progress evaluation protocols rely heavily on rank correlation along successful trajectories, where predicting linear frame index ($t/T$) trivially achieves near perfect correlation while completely masking failure blindness~\cite{leng2026flora}. To establish an unbiased, contamination free testbed, we construct \bench, spanning \textbf{90 unseen OOD manipulation tasks}, \textbf{1{,}101 trajectories}, and \textbf{31{,}088 densely annotated frames} strictly disjoint from all training splits. Every frame in \bench is annotated via our two-stage rubric annotation protocol (Sec.~\ref{sec:data_engine}) and verified or corrected by human reviewers so ground truth reflects true physical task accomplishment at every step. Crucially, partitioned into three coarse terminal progress tiers to align outcome discrimination across heterogeneous models, \bench maintains a balanced trajectory distribution across outcomes: only 442 trajectories (40.1\%) fall into the successful tier ($p^{\mathrm{GT}}_T \ge 2/3$), whereas 659 trajectories (59.9\%) are unsuccessful, comprising 110 suboptimal rollouts ($1/3 \le p^{\mathrm{GT}}_T < 2/3$) and 549 failed rollouts ($p^{\mathrm{GT}}_T < 1/3$). The benchmark integrates three complementary diagnostic subsets:
(1)~\emph{Real World OOD} (502 full length 32 frame trajectories across 40 unseen real robot tasks spanning Franka, xArm, WidowX/Trossen, Koch, and SO-101 platforms);
(2)~\emph{Truncated Prefixes} (259 incomplete 16 frame prefixes truncated midway from successful episodes across the same 40 tasks, directly testing whether models falsely predict 100\% at the end of any video clip); and
(3)~\emph{RoboFAC}~\cite{lu2025robofac} (340 detailed 32 frame trajectories across 50 tasks featuring diverse manipulation failure modes).

\textbf{Evaluation metrics.}
To evaluate local temporal ordering, global calibration across trajectories, terminal outcome discrimination, and pointwise precision simultaneously, \bench randomly samples 58{,}272 fixed frame pairs: 32 pairs per trajectory across all 1{,}101 trajectories (35{,}232 within trajectory pairs across Success and Unsuccessful rollouts) and 256 pairs per task reference group across all 90 tasks (23{,}040 cross trajectory pairs partitioned into Success--Success $\mathrm{S{\leftrightarrow}S}$, Unsuccessful--Unsuccessful $\mathrm{N{\leftrightarrow}N}$, and Success--Unsuccessful $\mathrm{S{\leftrightarrow}N}$), swapping each sampled pair $(a, b) \leftrightarrow (b, a)$ with $50\%$ probability so forward and backward orderings are uniformly distributed. Let $\mathrm{rel}(u, v; \epsilon) \in \{-1, 0, +1\}$ denote the ternary ordering relation with tie tolerance $\epsilon = 0.10$, defined as $0$ ($\approx$) if $|u - v| \le \epsilon$, $+1$ ($>$) if $u - v > \epsilon$, and $-1$ ($<$) if $u - v < -\epsilon$. We report five metrics:
(1)~\textbf{Within trajectory agreement ($\uparrow$)}: percentage of frame pairs $(a, b)$ from the same trajectory satisfying $\mathrm{rel}(\hat{p}_a, \hat{p}_b; 0.1) = \mathrm{rel}(p^{\mathrm{GT}}_a, p^{\mathrm{GT}}_b; 0.1)$;
(2)~\textbf{Cross trajectory agreement ($\uparrow$)}: percentage of frame pairs $(a, b)$ from distinct rollouts of the same task satisfying the same ternary agreement test;
(3)~\textbf{Terminal agreement ($\uparrow$)}: because continuous progress and reward models do not uniformly emit discrete success or failure labels, we adopt coarse three-bin calibration ($[0, 1/3)$ Failure, $[1/3, 2/3)$ Suboptimal, $[2/3, 1]$ Success) and measure the percentage of the 1{,}101 trajectories whose predicted final progress $\hat{p}_T$ falls into the same tier as $p^{\mathrm{GT}}_T$ to capture cross-model consistency in task outcome discrimination;
(4)~\textbf{Progress MAE ($\downarrow$)}: frame weighted mean absolute error $100 \times \frac{1}{N_{\mathrm{f}}}\sum |\hat{p} - p^{\mathrm{GT}}|$ in percentage points across all 31{,}088 frames with zero tie tolerance; and
(5)~\textbf{Average score ($\uparrow$)}: the unweighted arithmetic mean of Overall Within agreement, Overall Cross agreement, and Terminal agreement.

\textbf{Baselines and evaluation protocol.}
We benchmark \methodfour (step 3{,}000) and \methodtwentyseven (LoRA step 1{,}500) against 15 configurations spanning nine competitive model families:
(1)~the frontier $>10\text{T}$-parameter proprietary multimodal model \textbf{GPT-6-Astra} evaluated with high/max reasoning effort;
(2)~recent generalist open reward and process reward models, including \textbf{Flora Rewarder} (27B)~\cite{leng2026flora}, \textbf{Robometer-4B}~\cite{robometer2026} (evaluated under both 8 frame global uniform sampling with 32 frame interpolation and dense incremental prefix sampling), \textbf{Robo-Dopamine GRM} (3B and 8B) and \textbf{GRM 2.0 Preview} (4B and 8B)~\cite{tan2025robodopamine}, \textbf{RoboReward} (4B and 8B)~\cite{lee2026roboreward}, \textbf{NimCritic}~\cite{xu2026nimcritic}, and \textbf{DenseReward}~\cite{fang2026densereward} (single frame and 3 frame thinking); and
(3)~temporal distance value models \textbf{RynnValue} (4B and 8B)~\cite{huang2026rynnvalue}, which predict remaining seconds to completion (for RynnValue$^*$, negated remaining time with a 0.1\,s tie margin is used for pairwise ordering, while Terminal agreement and normalized $[0,1]$ MAE are omitted because uncalibrated remaining seconds do not map uniquely to completion ratios). For \method, each trajectory is evaluated frame by frame conditioned on the fixed reference keyframes and progress anchors of each task without temporal smoothing or endpoint forcing. Regarding baseline context conditioning, \textbf{GPT-6-Astra} is evaluated with the exact same interleaved reference keyframe and anchor prompt as \method; \textbf{GRM} and \textbf{GRM 2.0 Preview}, which natively accept boundary visual context, are provided with the initial and terminal frames extracted from the same reference trajectory; whereas text-driven reward baselines (\textbf{Flora Rewarder}, \textbf{Robometer-4B}, \textbf{RoboReward}, \textbf{NimCritic}, \textbf{DenseReward}, and \textbf{RynnValue}) do not support reference frame inputs and are trained for language-conditioned reward estimation, making native instruction-conditioned evaluation the direct comparison protocol. Complete training hyperparameters and evaluation details are provided in Appendix~\ref{sec:app_impl}.

\begin{figure}[!tp]
\centering
\includegraphics[width=\textwidth]{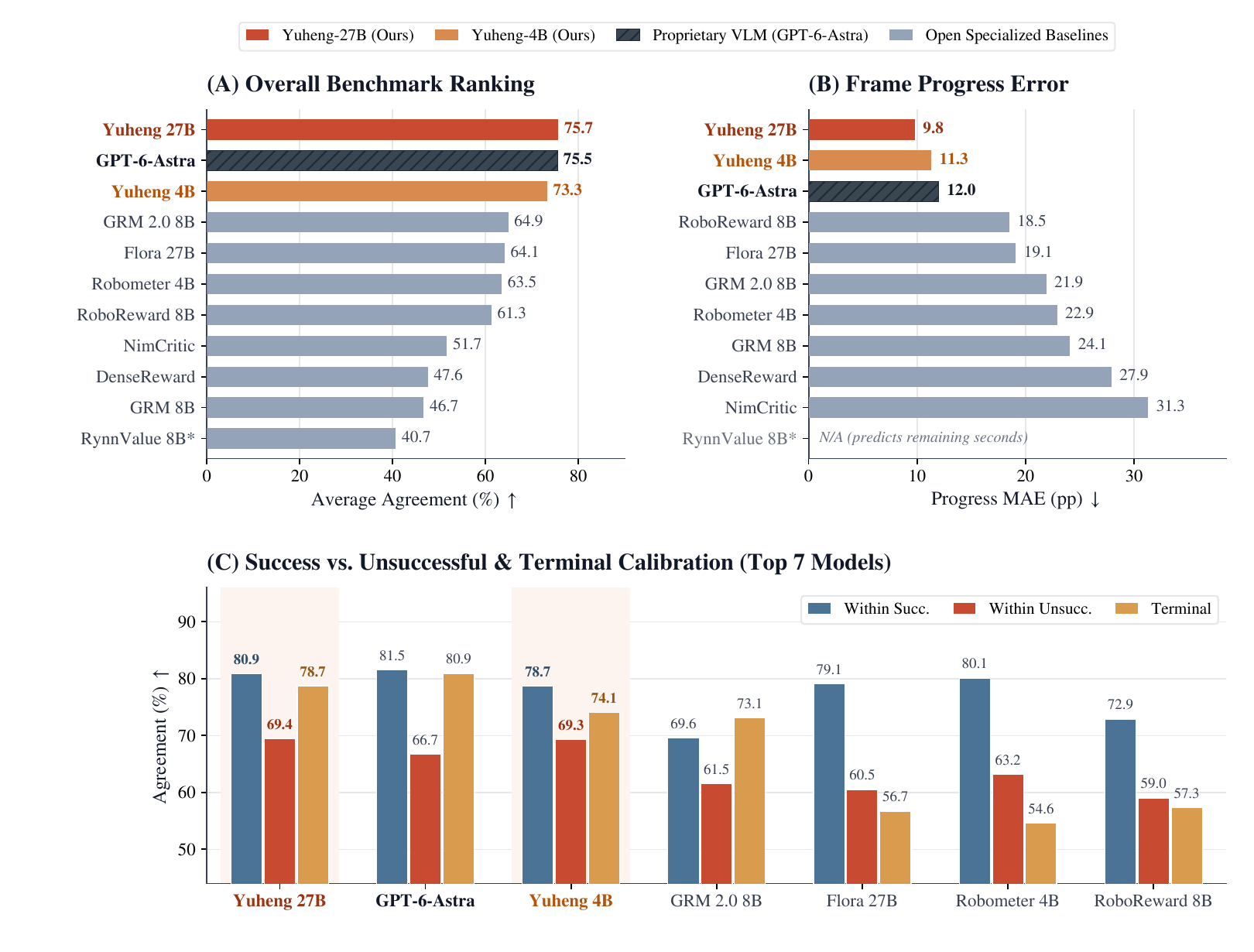}
\vspace{-6pt}
\caption{\textbf{Quantitative evaluation on \bench across 90 unseen manipulation tasks and 1{,}101 trajectories.}
(\textbf{A})~Overall Average score ranking (best configuration per model family). \methodtwentyseven (75.7\%) and \methodfour (73.3\%) outperform the strongest prior open baseline (GRM 2.0 8B, 64.9\%) by +10.8 pp and +8.4 pp, respectively, surpass all other calibrated open progress baselines by up to +29.0 pp, and exceed the temporal distance model RynnValue by up to +35.0 pp, with \methodtwentyseven edging past proprietary GPT-6-Astra (75.5\%).
(\textbf{B})~Dense Progress MAE in percentage points (lower is better), where \methodtwentyseven (9.8) and \methodfour (11.3) improve over GPT-6-Astra (12.0) and cut prior specialized baseline errors (18.5--31.3) nearly in half.
(\textbf{C})~Breakdown of Within Success agreement, Within Unsuccessful agreement, and three way Terminal agreement across the top 7 models. Whereas leading unanchored reward models such as Robometer-4B and Flora Rewarder drop sharply by 17 to 25 pp on unsuccessful rollouts and terminal classification, \method sustains strong calibration across failed and suboptimal executions. $^*$RynnValue predicts remaining time in seconds (MAE N/A).}
\label{fig:benchmark_overview}
\end{figure}

\begin{table}[!tp]
\centering
\caption{\textbf{Complete OOD benchmark results on \bench (90 tasks, 1{,}101 trajectories, 31{,}088 frames).} Agreement and Average scores are reported in percentages ($\uparrow$); Progress MAE is reported in percentage points ($\downarrow$). Unsuccessful ($\mathrm{N}$) aggregates suboptimal and failed rollouts; $\mathrm{S{\leftrightarrow}S}$, $\mathrm{N{\leftrightarrow}N}$, and $\mathrm{S{\leftrightarrow}N}$ denote cross trajectory outcome splits. \textbf{Bold} marks the best result per column (ranked at full precision prior to rounding). $^*$RynnValue predicts remaining time in seconds; its Average is computed over Within and Cross Overall agreement.}
\label{tab:main_benchmark}
\vspace{2pt}
\resizebox{\textwidth}{!}{%
\begin{tabular}{llcccccccccccc}
\toprule
\multirow{2}{*}{\textbf{Model}} & \multirow{2}{*}{\textbf{Configuration}} & \multirow{2}{*}{\textbf{Avg} $\uparrow$} & \multicolumn{3}{c}{\textbf{Within Agreement} $\uparrow$} & \multicolumn{4}{c}{\textbf{Cross Agreement} $\uparrow$} & \multirow{2}{*}{\textbf{Term.} $\uparrow$} & \multicolumn{3}{c}{\textbf{Progress MAE (pp)} $\downarrow$} \\
\cmidrule(lr){4-6}\cmidrule(lr){7-10}\cmidrule(lr){12-14}
 &  &  & \textbf{Overall} & \textbf{Succ} & \textbf{Unsucc} & \textbf{Overall} & $\mathbf{S{\leftrightarrow}S}$ & $\mathbf{N{\leftrightarrow}N}$ & $\mathbf{S{\leftrightarrow}N}$ &  & \textbf{Overall} & \textbf{Succ} & \textbf{Unsucc} \\
\midrule
\multicolumn{14}{@{}l}{\textit{Proprietary Frontier Vision--Language Model}} \\
GPT-6-Astra & Reasoning (high/max) & 75.5 & 72.7 & \textbf{81.5} & 66.7 & 72.9 & 77.8 & 66.0 & 75.9 & \textbf{80.9} & 12.0 & 9.4 & 14.1 \\
\midrule
\multicolumn{14}{@{}l}{\textit{Open Specialized Reward, Value, and Progress Models}} \\
GRM 2.0 Preview~\cite{tan2025robodopamine} & 8B & 64.9 & 64.8 & 69.6 & 61.5 & 56.8 & 66.1 & 50.7 & 58.3 & 73.1 & 21.9 & 17.8 & 25.2 \\
GRM 2.0 Preview~\cite{tan2025robodopamine} & 4B & 63.2 & 63.1 & 68.3 & 59.6 & 56.9 & 65.8 & 50.4 & 58.7 & 69.5 & 21.2 & 16.9 & 24.6 \\
Flora Rewarder~\cite{leng2026flora} & 27B & 64.1 & 68.0 & 79.1 & 60.5 & 67.5 & 78.7 & 58.7 & 70.1 & 56.7 & 19.1 & 14.5 & 22.7 \\
Robometer~\cite{robometer2026} & 4B (Global 8 frame interp.) & 63.5 & 70.0 & 80.1 & 63.2 & 66.0 & 76.1 & 58.2 & 68.2 & 54.6 & 22.9 & 18.9 & 26.2 \\
Robometer~\cite{robometer2026} & 4B (Incremental prefix) & 59.5 & 64.7 & 71.8 & 59.9 & 60.4 & 63.5 & 56.1 & 62.2 & 53.5 & 29.3 & 24.9 & 32.8 \\
RoboReward~\cite{lee2026roboreward} & 8B & 61.3 & 64.6 & 72.9 & 59.0 & 62.0 & 70.8 & 56.2 & 63.3 & 57.3 & 18.5 & 15.5 & 20.9 \\
RoboReward~\cite{lee2026roboreward} & 4B & 60.3 & 64.2 & 73.5 & 58.0 & 60.5 & 67.7 & 53.8 & 62.8 & 56.0 & 20.6 & 16.6 & 23.8 \\
NimCritic~\cite{xu2026nimcritic} & Default & 51.7 & 61.4 & 70.6 & 55.2 & 51.3 & 59.0 & 44.9 & 53.2 & 42.4 & 31.3 & 24.4 & 36.8 \\
DenseReward~\cite{fang2026densereward} & Single frame & 47.6 & 50.0 & 50.5 & 49.7 & 48.2 & 46.8 & 48.0 & 48.7 & 44.5 & 27.9 & 27.7 & 28.0 \\
DenseReward~\cite{fang2026densereward} & 3 frame thinking & 47.4 & 51.5 & 52.7 & 50.7 & 50.6 & 52.6 & 47.9 & 51.9 & 40.1 & 27.4 & 25.6 & 28.8 \\
GRM~\cite{tan2025robodopamine} & 8B & 46.7 & 49.8 & 39.7 & 56.5 & 39.8 & 35.4 & 47.7 & 36.0 & 50.5 & 24.1 & 32.7 & 17.3 \\
GRM~\cite{tan2025robodopamine} & 3B & 45.4 & 49.0 & 39.0 & 55.7 & 36.5 & 31.3 & 46.0 & 31.9 & 50.7 & 25.0 & 33.5 & 18.2 \\
RynnValue$^*$~\cite{huang2026rynnvalue} & 8B & 40.7 & 36.9 & 53.6 & 25.7 & 44.5 & 49.6 & 34.8 & 49.2 & -- & -- & -- & -- \\
RynnValue$^*$~\cite{huang2026rynnvalue} & 4B & 33.9 & 30.0 & 43.0 & 21.3 & 37.9 & 41.5 & 29.4 & 42.2 & -- & -- & -- & -- \\
\midrule
\multicolumn{14}{@{}l}{\textit{In-Context Semantic Progress Models (Ours)}} \\
\rowcolor{yuhengBg} \textbf{Yuheng (Ours)} & \textbf{4B (Qwen3-VL-4B)} & 73.3 & 73.1 & 78.7 & 69.3 & 72.8 & 77.9 & 66.9 & 75.1 & 74.1 & 11.3 & 11.1 & 11.4 \\
\rowcolor{yuhengBg} \textbf{Yuheng (Ours)} & \textbf{27B (Qwen3.8-27B LoRA)} & \textbf{75.7} & \textbf{74.0} & 80.9 & \textbf{69.4} & \textbf{74.2} & \textbf{79.0} & \textbf{67.3} & \textbf{77.3} & 78.7 & \textbf{9.8} & \textbf{9.3} & \textbf{10.2} \\
\bottomrule
\end{tabular}%
}
\end{table}

\subsection{State-of-the-Art Comparison on \bench}\label{sec:main_results}

Figure~\ref{fig:benchmark_overview} and Table~\ref{tab:main_benchmark} report comprehensive results across all 17 evaluated model configurations. \methodtwentyseven achieves the highest overall Average score of \textbf{75.7\%} and the lowest overall Progress MAE of \textbf{9.8} pp, ranking first in 10 of the 12 numerical columns in Table~\ref{tab:main_benchmark}. Compared with the strongest external specialized baseline, GRM 2.0 Preview 8B (64.9\% Average, 21.9 MAE), \methodtwentyseven improves Average agreement by \textbf{+10.8 pp}, boosts Cross agreement by \textbf{+17.4 pp} (74.2\% vs.\ 56.8\%), and cuts frame error by \textbf{55\%} (9.8 vs.\ 21.9).

\textbf{Controlled backbone comparisons against Robometer and Flora Rewarder.}
Comparing against specialized baselines from the same VLM families isolates the impact of \method's formulation and data engine from backbone capacity. First, \methodfour and Robometer-4B~\cite{robometer2026} share the exact same Qwen3-VL-4B backbone. Despite training on 65 times fewer trajectories than Robometer-4B (16{,}321 vs.\ 1{,}059{,}370), \methodfour improves Average score by \textbf{+9.8 pp} over Robometer's best global sampling configuration (73.3\% vs.\ 63.5\%) and by \textbf{+13.8 pp} over Robometer's causal incremental prefix mode (73.3\% vs.\ 59.5\%), while slashing MAE from 22.9 and 29.3 down to \textbf{11.3}. Second, at the 27B hybrid Qwen scale, \methodtwentyseven (adapting only 0.65\% LoRA and head parameters on Qwen3.8-27B~\cite{qwen38} with 16{,}321 trajectories) outperforms full model Flora Rewarder~\cite{leng2026flora} (built on Qwen3.5-27B~\cite{qwen35}) by \textbf{+11.6 pp} in Average score (75.7\% vs.\ 64.1\%), \textbf{+22.0 pp} in Terminal agreement (78.7\% vs.\ 56.7\%), and reduces MAE from 19.1 to \textbf{9.8}; even the smaller \methodfour surpasses 27B Flora Rewarder by \textbf{+9.2 pp} in Average score (73.3\% vs.\ 64.1\%).

\textbf{Eliminating failure blindness and surpassing proprietary GPT-6-Astra.}
Inspecting the Success versus Unsuccessful splits in Fig.~\ref{fig:benchmark_overview}C explains why prior reward models falter on realistic rollouts. On successful trajectories alone, Robometer-4B (80.1\%) and Flora Rewarder (79.1\%) attain competitive Within ordering because monotonic time correlates strongly with progress when nothing goes wrong. On unsuccessful rollouts, however, Robometer-4B plunges by 16.9 points to 63.2\% (and 54.6\% Terminal agreement) and Flora Rewarder drops by 18.6 points to 60.5\% (56.7\% Terminal agreement), as both overpredict completion on stalled attempts and truncated prefixes (achieving only 58.0\% and 57.9\% Average on the \emph{Truncated Prefixes} subset, compared with 74.7\% for \methodtwentyseven; see Appendix Table~\ref{tab:subset_breakdown}). Notably, even though the $>10\text{T}$-parameter proprietary model GPT-6-Astra receives the exact same reference keyframe prompt with high/max reasoning effort, our 27B model (updating only 170.6M parameters in a single forward pass) edges it out in Overall Average (75.7\% vs.\ 75.5\%), while both \methodtwentyseven and \methodfour outperform it on unsuccessful trajectories across Within agreement (69.4\% and 69.3\% vs.\ 66.7\%), Cross $\mathrm{N{\leftrightarrow}N}$ agreement (67.3\% and 66.9\% vs.\ 66.0\%), and Progress MAE (10.2 and 11.4 vs.\ 14.1 pp), demonstrating that dedicated semantic alignment yields tighter failure calibration than massive in-context frontier reasoning at a tiny fraction of parameter scale and inference cost.

\subsection{Qualitative Progress Trajectory Analysis}\label{sec:qualitative}

\begin{figure}[!tp]
\centering
\includegraphics[width=\textwidth]{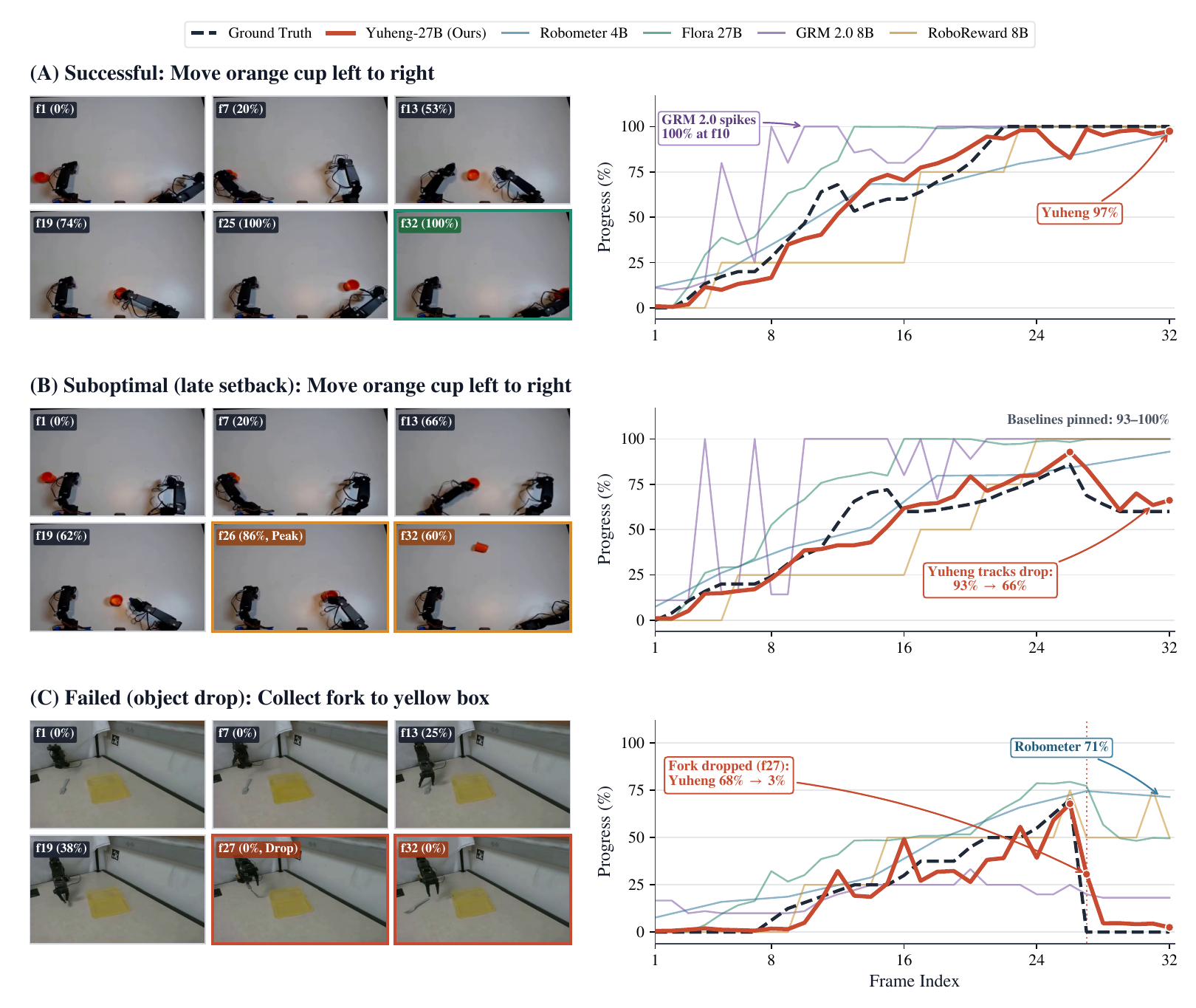}
\vspace{-6pt}
\caption{\textbf{Qualitative comparison of frame by frame progress predictions on unseen \bench rollouts.}
(\textbf{A})~On a successful query execution of \emph{Move the orange cup from left to right}, \methodtwentyseven tracks the semantic ground truth smoothly from 0\% to 97\% without premature saturation, whereas GRM 2.0 8B spikes erratically to 100\% at frame 10.
(\textbf{B})~On a suboptimal rollout of the same task where the cup is displaced near the end (ground truth peaks at 86\% at frame 26 and regresses to 60\%), \methodtwentyseven captures both the peak and the terminal drop to 66\%, while Flora Rewarder, RoboReward 8B, and Robometer 4B remain saturated between 93\% and 100\%.
(\textbf{C})~On an irreversible failure of \emph{Collect the fork to the yellow box} where the robot lifts the fork (70\% at frame 26) and then drops it outside the box (0\% at frame 27), \methodtwentyseven immediately drops from 68\% back to 3\%, whereas Robometer 4B (71\%) and Flora Rewarder (50\%) fail to detect the undone progress.}
\label{fig:qualitative_curves}
\end{figure}

To illustrate how quantitative gains manifest along complete episodes, Fig.~\ref{fig:qualitative_curves} compares raw, unsmoothed 32 frame predictions from \methodtwentyseven against four leading specialized baselines across three diagnostic rollouts on \bench. During successful manipulation (Fig.~\ref{fig:qualitative_curves}A), GRM 2.0 Preview 8B and Flora Rewarder saturate prematurely at 100\% halfway through transport (frames 10--14), leaving zero reward gradient for final placement, whereas \methodtwentyseven rises in lockstep with physical cup displacement. When execution experiences a late perturbation that degrades final placement quality (Fig.~\ref{fig:qualitative_curves}B), baseline predictions remain pinned near 100\% because they anchor heavily on elapsed episode length; in contrast, \methodtwentyseven tracks the reversal from 93\% at frame 26 down to 66\% at termination. Most strikingly, when an object slip undoes prior manipulation entirely (Fig.~\ref{fig:qualitative_curves}C, dropping the lifted fork outside the target container at frame 27), \methodtwentyseven drops precipitously from 68\% to 3\% within two frames, closely mirroring ground truth, whereas Robometer-4B continues climbing to 71\% and Flora Rewarder plateaus at 50\%.

\subsection{Core Capabilities of Universal Progress Modeling}\label{sec:capabilities}

Beyond standard benchmark accuracy, conditioning explicitly on labeled reference keyframes endows \method with four practical capabilities that unanchored reward models cannot support.

\begin{figure}[!tp]
\centering
\includegraphics[width=\textwidth]{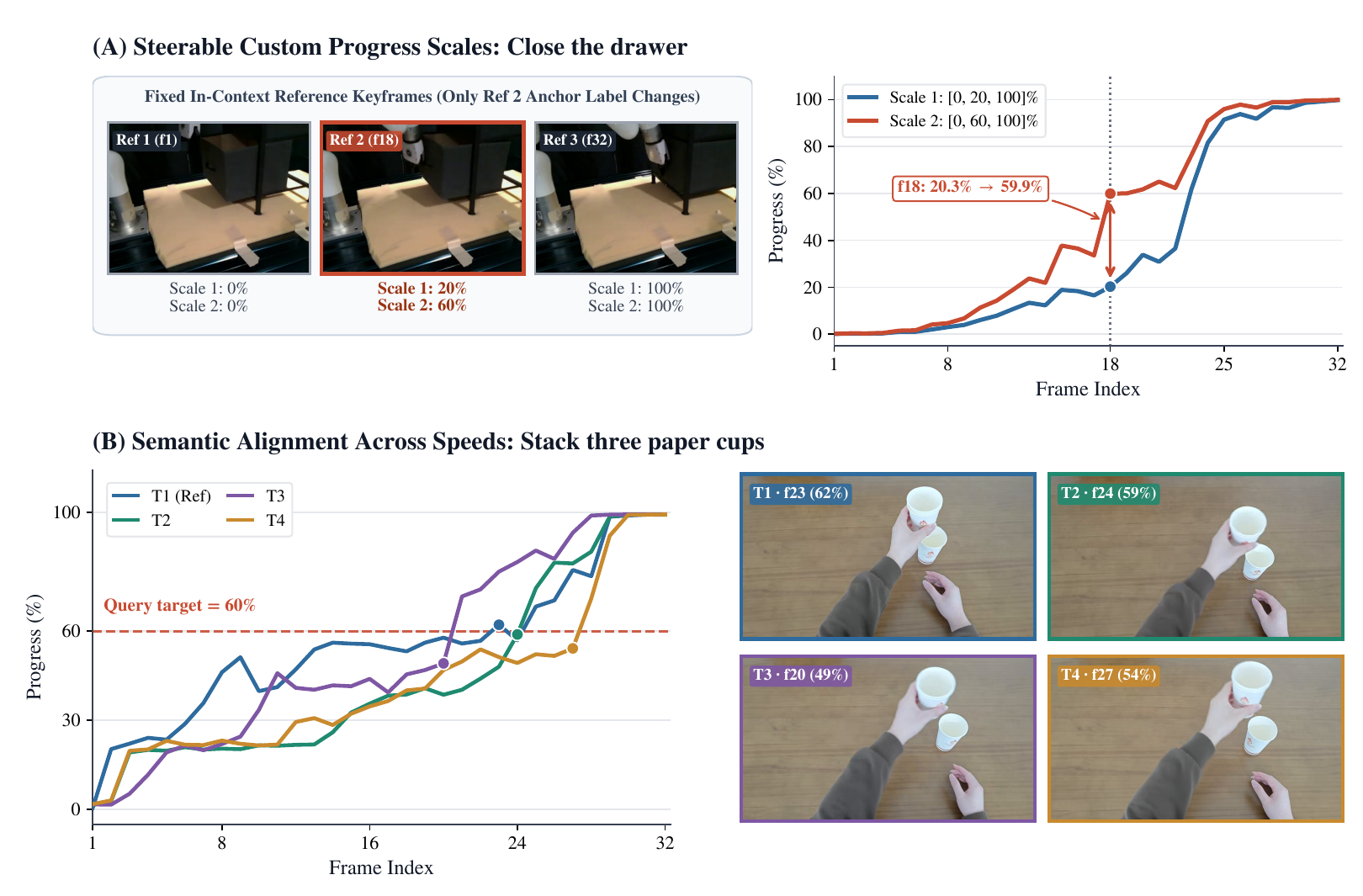}
\vspace{-6pt}
\caption{\textbf{Steerable custom progress scales and semantic alignment across trajectories.}
(\textbf{A})~Given identical reference keyframes (frames 1, 18, and 32 of \emph{Close the drawer}), changing only the prompt anchor value of Keyframe 2 from 20\% (Scale 1) to 60\% (Scale 2) steers \methodtwentyseven's prediction at frame 18 from 20.3\% to 59.9\% without parameter updates, smoothing, or endpoint clamping, reshaping the entire trajectory curve to follow the user's rubric.
(\textbf{B})~Conditioned on reference keyframes from \texttt{T1} (\texttt{T1 (Ref)}) of \emph{Stack the three paper cups in order from left to right}, \methodtwentyseven aligns three unseen variable speed executions (\texttt{T2}--\texttt{T4}) onto the same semantic progress coordinate system; selecting the discrete frame closest to the 60\% target progress threshold retrieves the exact stage where the middle cup is poised over the rightmost cup across \texttt{T1} and all three held-out executions (frames 23, 24, 20, and 27, with predicted scores spanning 49\%--62\% under 32 frame sampling).}
\label{fig:scale_and_align}
\end{figure}

\textbf{Steerable custom progress scales.}
Different downstream consumers often require distinct reward shaping profiles for the same physical task: a precision insertion controller may allocate 20\% progress to coarse approach and 80\% to contact alignment, whereas a reach curriculum may assign 60\% to reaching the approach pose. Figure~\ref{fig:scale_and_align}A tests whether \methodtwentyseven genuinely obeys arbitrary custom scales at inference time on \emph{Close the drawer}. Holding the three reference images (frames 1, 18, and 32) and all query frames fixed while changing only the prompt anchor specification from $[0\%, 20\%, 100\%]$ to $[0\%, 60\%, 100\%]$, \methodtwentyseven shifts its unsmoothed prediction at frame 18 from \textbf{20.3\%} to \textbf{59.9\%} (while keeping frame 1 at 0.1--0.2\% and frame 32 at 99.8--99.9\%) and smoothly recalibrates every intermediate frame before and after the milestone. This confirms that \method treats in-context progress labels as an active metric specification rather than memorizing a single canonical curve per task.

\textbf{Semantic progress alignment across trajectories.}
Because \method scores query states relative to visual milestones rather than frame indices, trajectories executed at disparate speeds map onto a shared semantic coordinate system. In Fig.~\ref{fig:scale_and_align}B, conditioned on reference keyframes drawn from \texttt{T1}, \methodtwentyseven aligns three unseen variable speed human demonstrations (\texttt{T2}--\texttt{T4}) of stacking three paper cups from left to right despite pronounced temporal variation (reaching the second stacking phase anywhere between frame 20 and frame 27). Selecting the discrete frame closest to a target progress of 60\% along each predicted curve retrieves visually homologous semantic states across both \texttt{T1} and all three query rollouts (frames 23, 24, 20, and 27, with predicted progress values of 62\%, 59\%, 49\%, and 54\% across sharp stage transitions under 32 frame sampling, each capturing the hand positioning the middle stack above the rightmost cup), directly enabling automated keyframe transfer across unannotated datasets.

\begin{figure}[!tp]
\centering
\includegraphics[width=\textwidth]{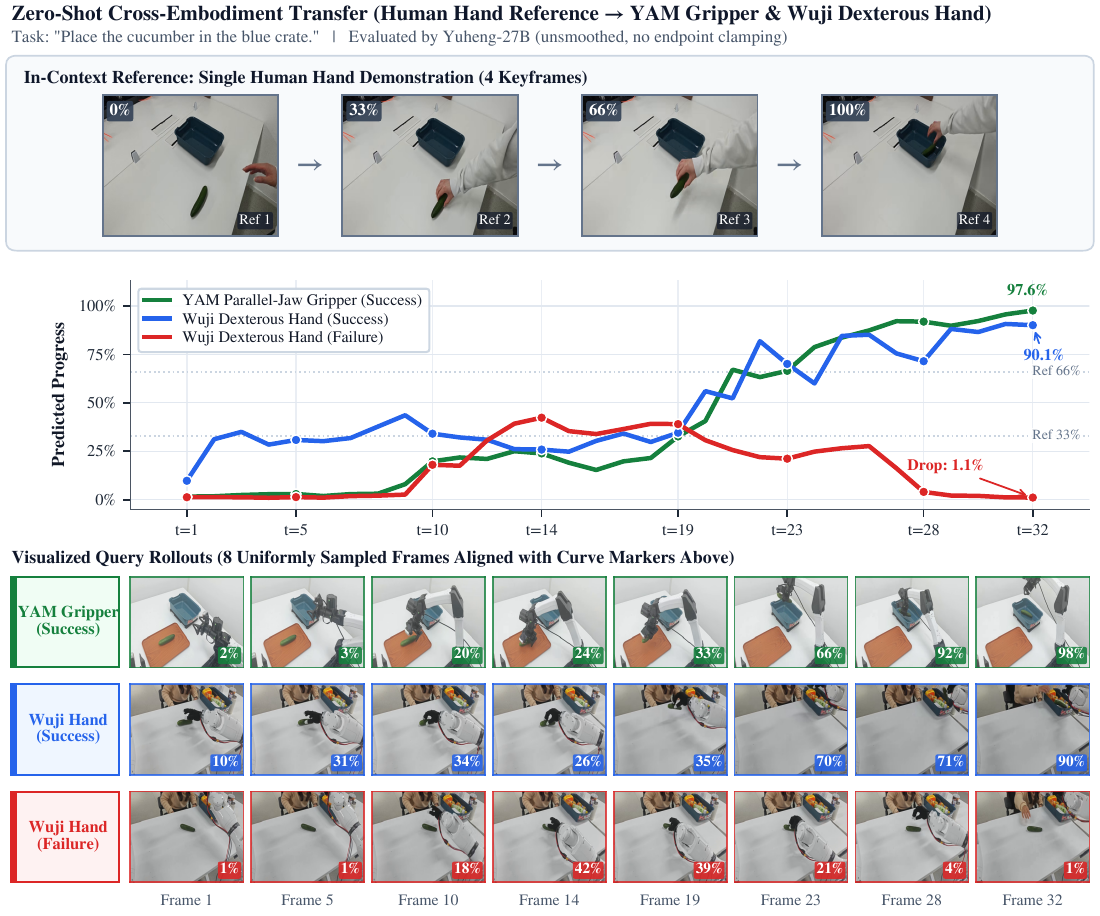}
\vspace{-6pt}
\caption{\textbf{Zero-shot transfer across embodiments from a single human video.}
\emph{Top:} Four reference keyframes ($[0\%, 33\%, 66\%, 100\%]$) extracted from a single \emph{human hand} demonstration of \emph{Place the cucumber in the blue crate} serve as the sole in-context progress scale.
\emph{Middle:} Without task-specific retraining, temporal smoothing, or endpoint clamping, \methodtwentyseven transfers this human reference directly to a YAM parallel jaw robotic arm (Success, climbing steadily from 1.6\% to 97.6\%), a five finger Wuji dexterous hand (Success, reaching 90.1\%), and a failed Wuji dexterous trial where the cucumber slips outside the crate (peaking at 42.3\% during grasp at frame 14 before collapsing to 1.1\% after the drop).
\emph{Bottom:} Eight uniformly sampled frames (frames 1, 5, 10, 14, 19, 23, 28, and 32) from each rollout aligned vertically with the curve markers above, annotated with the predicted progress score at every frame.}
\label{fig:cross_embodiment}
\end{figure}

\textbf{Zero-shot transfer across embodiments from human video.}
Collecting successful robot demonstrations to define a reference scale is undesirable when bootstrapping a new robot morphology or task from scratch, whereas recording a single human demonstration takes seconds~\cite{chen2026host,EgoScale}. Figure~\ref{fig:cross_embodiment} evaluates whether \methodtwentyseven can bridge large morphological and viewpoint gaps using a single human hand video of \emph{Place the cucumber in the blue crate} (four keyframes labeled $[0\%, 33\%, 66\%, 100\%]$) as the sole in-context reference. Conditioned solely on these four human hand keyframes at inference time, \methodtwentyseven transfers the reference scale seamlessly to both a YAM parallel jaw gripper arm (climbing steadily to 97.6\% upon placement) and a five finger Wuji dexterous robotic hand (climbing to 90.1\% on a successful trial), while sharply penalizing a failed Wuji dexterous trial where the cucumber is grasped (peaking at 42.3\% at frame 14) but subsequently dropped onto the table (collapsing to 1.1\% by termination). The aligned frame sequences beneath the curves confirm that each rise, plateau, and drop tracks physical object state rather than embodiment appearance or elapsed frame index.

\textbf{One-shot generalization across manipulation domains.}
We further evaluate \methodtwentyseven across four distinct manipulation domains (Simulation, Gripper Manipulation, Human Manipulation, and Dexterous Manipulation), conditioning each unseen task on a single successful demonstration. Appendix Fig.~\ref{fig:app_oneshot} (Appendix~\ref{sec:supp_oneshot}) presents paired successful and failed trajectories from all four domains with eight sampled frames aligned above each full-width progress curve. Across simulated mug rearrangement and plug insertion, real parallel jaw block sorting and slipper placement, egocentric Rubik's Cube rotation and cup stacking, and five finger bimanual drink handoffs and snack extraction, \methodtwentyseven consistently produces smooth monotonic curves approaching 100\% on successful executions while assigning bounded partial progress or sharp regressions when attempts stall or fail.

\subsection{Mechanistic Ablations and Scaling Analysis}\label{sec:ablations}

\begin{figure}[!tp]
\centering
\includegraphics[width=\textwidth]{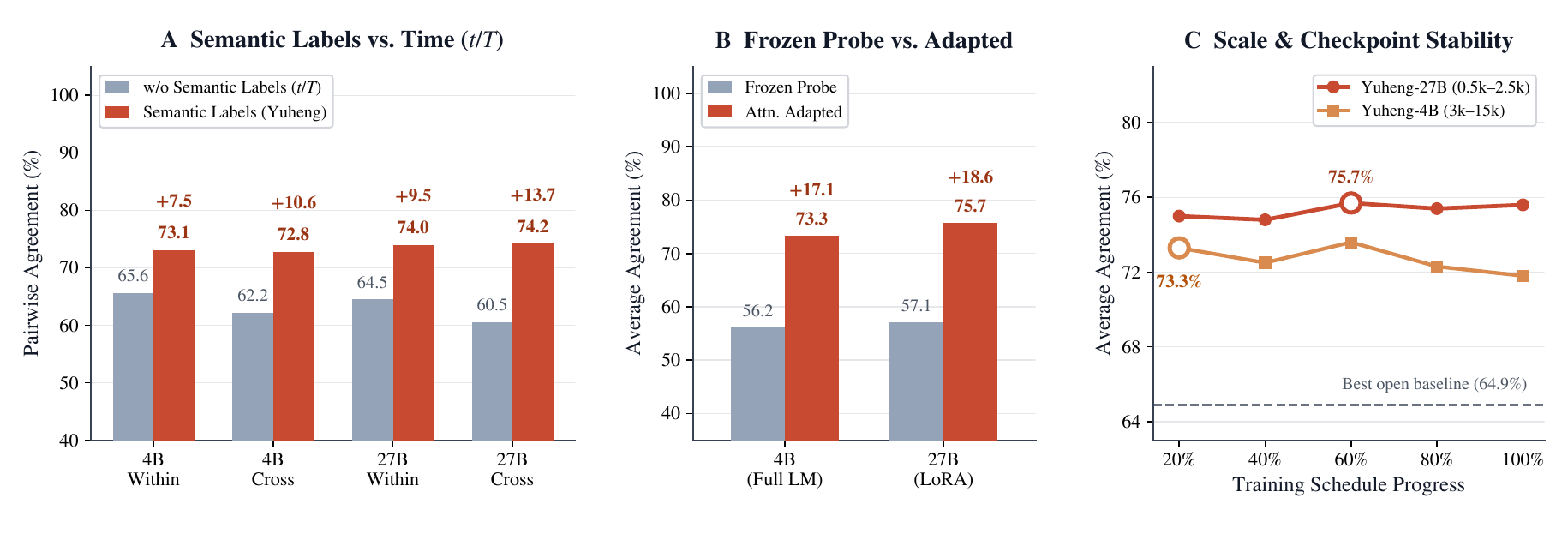}
\vspace{-6pt}
\caption{\textbf{Mechanistic ablations of supervision, backbone adaptation, and model scale on \bench (1{,}101 trajectories, 58{,}272 pairs).}
(\textbf{A})~Holding both the \method in-context architecture and backbone adaptation fixed on the full \bench, replacing same-trajectory elapsed-time ($t/T$) interpolation (\emph{Yuheng w/o Semantic Labels}) with cross-trajectory calibrated semantic supervision boosts Within agreement by +7.5 and +9.5 pp, Cross agreement by +10.6 and +13.7 pp, Terminal agreement by +13.8 and +28.1 pp, and Average score by +10.6 and +17.2 pp on 4B and 27B backbones, respectively, while cutting Progress MAE by 48\%--60\%.
(\textbf{B})~Training only the bilinear scale head over a frozen VLM (\emph{Frozen Backbone Probe}) degrades Average agreement by 17.1 pp on 4B and 18.6 pp on 27B relative to multimodal attention adaptation.
(\textbf{C})~Evaluation across training checkpoints shows that \methodtwentyseven consistently exceeds \methodfour by +1.7 to +3.8 pp and surpasses all prior open baselines from step 500 onward.}
\label{fig:ablations}
\end{figure}

\begin{table}[!tp]
\centering
\caption{\textbf{Summary of controlled mechanistic ablations on the full 1{,}101-trajectory \bench (58{,}272 pairs).} Panel~A compares \emph{Yuheng w/o Semantic Labels} (trained with same-trajectory $t/T$ supervision) against cross-trajectory calibrated semantic supervision under the identical in-context architecture and backbone adaptation. Panel~B compares frozen-backbone probing against multimodal attention adaptation under identical semantic supervision. Progress MAE (pp, $\downarrow$) is reported in parentheses alongside Average agreement.}
\label{tab:ablations}
\vspace{2pt}
\resizebox{\textwidth}{!}{%
\begin{tabular}{llcccccc}
\toprule
\textbf{Ablation Study} & \textbf{Model Variant} & \textbf{Backbone Adaptation} & \textbf{Supervision} & \textbf{Within} $\uparrow$ & \textbf{Cross} $\uparrow$ & \textbf{Term.} $\uparrow$ & \textbf{Avg} $\uparrow$ \textbf{(MAE $\downarrow$)} \\
\midrule
\multicolumn{8}{@{}l}{\textit{A. Effect of Cross-Trajectory Semantic Labels vs.\ Same-Trajectory Time Supervision (\bench, 1{,}101 Trajectories)}} \\
Yuheng w/o Semantic Labels (4B) & Qwen3-VL-4B & Full LM + Head & Same-traj.\ time ($t/T$) & 65.6 & 62.2 & 60.3 & 62.7 (21.9) \\
\rowcolor{yuhengBg} \textbf{\methodfour} & Qwen3-VL-4B & Full LM + Head & \textbf{Cross-traj.\ semantic} & \textbf{73.1} & \textbf{72.8} & \textbf{74.1} & \textbf{73.3} (\textbf{11.3}) \\
Yuheng w/o Semantic Labels (27B) & Qwen3.8-27B & Full Attn LoRA + Head & Same-traj.\ time ($t/T$) & 64.5 & 60.5 & 50.6 & 58.5 (24.6) \\
\rowcolor{yuhengBg} \textbf{\methodtwentyseven} & Qwen3.8-27B & Full Attn LoRA + Head & \textbf{Cross-traj.\ semantic} & \textbf{74.0} & \textbf{74.2} & \textbf{78.7} & \textbf{75.7} (\textbf{9.8}) \\
\midrule
\multicolumn{8}{@{}l}{\textit{B. Effect of Multimodal Attention Adaptation vs.\ Frozen Backbone Probe (\bench, 1{,}101 Trajectories)}} \\
Frozen Backbone Probe (4B) & Qwen3-VL-4B & Frozen VLM (Head only) & Cross-traj.\ semantic & 63.4 & 58.9 & 46.3 & 56.2 (24.9) \\
\rowcolor{yuhengBg} \textbf{\methodfour} & Qwen3-VL-4B & \textbf{Full LM + Head} & Cross-traj.\ semantic & \textbf{73.1} & \textbf{72.8} & \textbf{74.1} & \textbf{73.3} (\textbf{11.3}) \\
Frozen Backbone Probe (27B) & Qwen3.8-27B & Frozen VLM (Head only) & Cross-traj.\ semantic & 64.4 & 63.0 & 43.9 & 57.1 (22.1) \\
\rowcolor{yuhengBg} \textbf{\methodtwentyseven} & Qwen3.8-27B & \textbf{Full Attn LoRA + Head} & Cross-traj.\ semantic & \textbf{74.0} & \textbf{74.2} & \textbf{78.7} & \textbf{75.7} (\textbf{9.8}) \\
\bottomrule
\end{tabular}%
}
\end{table}

To verify the individual contributions of our core design choices, we conduct three controlled ablation studies reported in Fig.~\ref{fig:ablations} and Table~\ref{tab:ablations}.

\textbf{Effect of calibrated semantic labels vs.\ same-trajectory time ($t/T$) supervision.}
A natural question is whether in-context keyframe conditioning alone suffices without semantic progress annotations. Without semantic labels, training cannot align query states across distinct trajectories or incorporate suboptimal and failed rollouts; it therefore degrades to sampling reference keyframes and queries from the same successful trajectory under linear frame-index ($t/T$) interpolation (\emph{Yuheng w/o Semantic Labels}). Evaluated on the full 1{,}101-trajectory \bench under the identical in-context architecture, backbone adaptation, tokenization, and protocol (Fig.~\ref{fig:ablations}A and Table~\ref{tab:ablations}A), \emph{Yuheng w/o Semantic Labels} reaches only 62.7\% Average agreement (21.9 pp MAE) on 4B and drops further to 58.5\% Average agreement (50.6\% Terminal agreement, 24.6 pp MAE) on 27B LoRA (step 1{,}000), as higher backbone capacity memorizes elapsed-time shortcuts more aggressively under temporal supervision. Introducing calibrated semantic progress labels with cross-trajectory, diverse-outcome query sampling lifts 4B performance across every dimension---Within agreement from 65.6\% to \textbf{73.1\% (+7.5 pp)}, Cross agreement from 62.2\% to \textbf{72.8\% (+10.6 pp)}, Terminal agreement from 60.3\% to \textbf{74.1\% (+13.8 pp)}, and Average score from 62.7\% to \textbf{73.3\% (+10.6 pp)} while cutting MAE from 21.9 to \textbf{11.3 pp ($-10.6$ pp)}. At 27B scale under the exact same LoRA adaptation, semantic supervision delivers even larger gains: Within rises from 64.5\% to \textbf{74.0\% (+9.5 pp)}, Cross from 60.5\% to \textbf{74.2\% (+13.7 pp)}, Terminal from 50.6\% to \textbf{78.7\% (+28.1 pp)}, and Average from 58.5\% to \textbf{75.7\% (+17.2 pp)}, slashing MAE by 60\% from 24.6 to \textbf{9.8 pp ($-14.8$ pp)}. Without semantic labels and cross-trajectory queries, the model estimates where a frame sits along typical video duration rather than whether the physical task state matches the reference anchors, causing terminal calibration and comparisons across successful and stalled rollouts to break down.

\textbf{Necessity of multimodal attention adaptation.}
To test whether standard VLM representations already encode progress alignment across images without backbone tuning, we train the projection head on the exact same semantic training split while keeping all VLM language and vision layers frozen (\emph{Frozen Backbone Probe}, Fig.~\ref{fig:ablations}B and Table~\ref{tab:ablations}B). On \bench, freezing Qwen3-VL-4B yields only 56.2\% Average (46.3\% Terminal agreement, 24.9 MAE at converged step 9{,}000; 50.4\% Average and 26.7 MAE at step 3{,}000), and freezing Qwen3.8-27B yields only 57.1\% Average (43.9\% Terminal agreement, 22.1 MAE). Adapting the language model in \methodfour (+17.1 Average points, +27.8 Terminal agreement points, $-13.6$ MAE) or exclusively the 16 full attention QKVO layers via LoRA in \methodtwentyseven (+18.6 Average points, +34.8 Terminal agreement points, $-12.3$ MAE) is essential to transform general visual--linguistic features into precise state correspondence across frames.

\textbf{Model scale, checkpoint stability, and data efficiency.}
Figure~\ref{fig:ablations}C tracks \bench Average performance across the full training schedule of both models (steps 3{,}000--15{,}000 for \methodfour; steps 500--2{,}500 for \methodtwentyseven; full checkpoint metrics in Appendix Table~\ref{tab:ckpt_sweep}), confirming that performance is robust to checkpoint choice rather than relying on test-set selection. Scaling from 4B to 27B yields consistent gains across all metrics (+2.4 Average points, +4.6 Terminal agreement points, and $-1.5$ MAE at the standard release checkpoints, step 3{,}000 for \methodfour and step 1{,}500 for \methodtwentyseven) while exhibiting remarkably rapid convergence: \methodtwentyseven already reaches 75.0\% Average at step 500 (surpassing every prior open baseline by over 10 points) and remains stable at 74.8\%--75.7\% through step 2{,}500 without late degradation. For full-LM \methodfour, the early-converged step 3{,}000 release checkpoint achieves strong calibration (73.3\% Average, 73.1\% Within agreement, 11.3 pp MAE); while extended updates fluctuate mildly between 71.8\% and 73.6\% Average (with Overall Within declining slightly and Progress MAE drifting to 11.4--12.0 pp), every saved checkpoint across both scales still outperforms all prior open baselines by 6.9 to 10.8 pp. Crucially, both models achieve these results from just \textbf{16{,}321} training trajectories, demonstrating that accurate semantic alignment supersedes brute force million scale temporal supervision.

%% file: chapters/applications.tex
\section{Downstream Robotic Applications}\label{sec:applications}

Intrinsic evaluation on \bench and cross-embodiment analyses (Sec.~\ref{sec:capabilities}) establish that \method tracks physical task accomplishment accurately across unseen tasks, mixed execution qualities, and diverse embodiments. Beyond offline evaluation, these verified calibration properties endow \method with broad application potential across the robot learning lifecycle. Below, we discuss five representative downstream paradigms spanning the four application pillars illustrated in Fig.~\ref{fig:teaser}D and highlight the practical advantages unlocked by its semantically anchored formulation.

\textbf{Online reinforcement refinement.}
In closed loop online RL atop pretrained visuomotor and vision--language--action (VLA) policies~\cite{DiffusionPolicy,pi0,pi05,OpenVLA}---such as latent-space Diffusion Steering RL (DSRL)~\cite{wagenmaker2025steering} and online VLA-RL~\cite{lu2025vlarlmasterfulgeneralrobotic,li2025simplevlarl}---sparse binary success signals and manual reward engineering or termination labeling are major bottlenecks. \method can serve as both a potential-based shaped reward provider ($r_t = \gamma \hat{p}_t - \hat{p}_{t-1} + r_{\mathrm{term}}\,\mathbb{I}[\hat{p}_t \ge \tau_{\mathrm{succ}}]$) and an autonomous episode completion detector. Conditioning on a single demonstration eliminates manual reward engineering for each task, while semantic progress calibration mitigates reward hacking such as lingering near target objects: because predicted progress plateaus during idling and drops immediately upon physical setbacks, stationary states yield non-positive shaped rewards and regressions incur immediate penalties. Moreover, \method's steerable reference scale lets practitioners emphasize critical manipulation stages without retraining, and the compact \methodfour backbone is well suited for low latency inference during online rollouts.

\textbf{Offline reinforcement refinement and quality weighted imitation.}
Real robot repositories inevitably mix expert demonstrations with hesitant, suboptimal, and failed trajectories. In advantage conditioned offline RL frameworks such as $\pi^*_{0.6}$ RECAP~\cite{intelligence2025pi06vla}, \method can act directly as a calibrated state value model for computing empirical action advantages; in standard imitation pipelines such as Advantage Weighted Behavior Cloning (AW-BC)~\cite{awr,kostrikov2022offline}, its local progress gains can supply sample weights to prioritize state advancing transitions. A key practical advantage over conventional offline refinement pipelines is that \method operates as a \emph{training free value estimator}: rather than training and tuning a separate value network for every task or dataset, practitioners can obtain calibrated values directly from a single reference demonstration, effectively suppressing or filtering out stalled and regressing frames while harvesting productive subsegments from imperfect rollouts.

\textbf{Data curation, teleoperation quality assessment, and collection standardization.}
As robotics datasets scale across teleoperated play logs, crowd evaluations, and autonomous fleets~\cite{OpenXEmbodiment,khazatsky2024droid,atreya2025roboarena}, ensuring demonstration quality both during data acquisition and prior to model training becomes critical for VLA pretraining and supervised fine-tuning~\cite{budzianowski2025opengvl,memmel2025strap,leng2026flora}. \method supports this data lifecycle from two complementary angles. First, \emph{during data collection}, an ordered reference keyframe sequence paired with milestone anchors establishes an explicit, standardized task specification, guiding operators toward consistent stage execution, supporting teleoperation quality assessment, and verifying demonstration completeness upon recording. Second, \emph{for pretraining and fine-tuning corpus filtering}, given that single reference demonstration per task group, \method can serve as an automated trajectory and segment quality scorer that identifies genuinely completed executions, penalizes rollouts containing repeated physical regressions, and filters out prolonged stalled or regressing frames without manual auditing or specialized quality classifiers.

\textbf{Runtime execution monitoring, failure detection, and autonomous stopping.}
Safe deployment of autonomous manipulation policies in unstructured environments requires identifying stalls and irreversible mistakes before episode timeouts or compounding manipulation errors occur~\cite{duan2024aha,lu2025robofac,dai2025racer}. Because \method evaluates incoming observations online, frame by frame, without future lookahead, it can operate directly as a lightweight runtime execution monitor: sustained progress plateaus expose missed grasps or kinematic jams, while sharp drops relative to prior peak progress immediately flag physical setbacks such as dropped objects or collapsed stacks (Fig.~\ref{fig:qualitative_curves}C). Unlike retrospective episode classifiers or temporal proxy estimators that remain falsely optimistic on stuck attempts, \method enables executive controllers to trigger autonomous task stopping once verified completion is reached, or initiate recovery routines and human intervention as soon as execution degrades.

\textbf{One-shot semantic keyframe transfer.}
Hierarchical visuomotor policies, subgoal planners~\cite{SuSIE,chen2026host}, and phase segmented manipulation datasets rely on consistent subtask boundaries across trajectories. Leveraging \method's semantic alignment across trajectories (Sec.~\ref{sec:capabilities}, Fig.~\ref{fig:scale_and_align}B), an annotator labels key semantic milestones once on a single source demonstration, and \method can automatically locate the matching stage frames across unannotated target rollouts of arbitrary speed by retrieving the discrete frame first reaching or closest to each target progress level. This eliminates manual stage labeling across an entire corpus, avoids spuriously assigning unreached stages to failed or truncated episodes, and transfers directly across embodiments---allowing milestones marked on a single video of a human hand to segment parallel jaw or dexterous robot datasets (Fig.~\ref{fig:cross_embodiment}).

%% file: chapters/appendix.tex
\begin{appendices}
\counterwithin{figure}{section}
\counterwithin{table}{section}

\section*{Appendix Contents}
\begin{enumerate}
  \item[A.] Implementation and training details\dotfill\pageref{sec:app_impl}
  \begin{enumerate}
    \item[A.1] Backbone architectures and projection head configuration\dotfill\pageref{sec:supp_arch}
    \item[A.2] Optimization, data mixture, and dynamic sampling protocol\dotfill\pageref{sec:supp_train}
    \item[A.3] Baseline evaluation protocol on \bench\dotfill\pageref{sec:supp_baselines}
  \end{enumerate}
  \item[B.] Interleaved multimodal prompt specification and token indexing\dotfill\pageref{sec:app_prompt}
  \item[C.] Extended benchmark breakdown, checkpoint sweep, and domain generalization\dotfill\pageref{sec:app_extra_eval}
  \begin{enumerate}
    \item[C.1] Subset evaluation across real world OOD, truncated prefixes, and RoboFAC\dotfill\pageref{sec:supp_subsets}
    \item[C.2] Complete training checkpoint sweep for \methodfour and \methodtwentyseven\dotfill\pageref{sec:supp_ckpts}
    \item[C.3] One-shot generalization across manipulation domains\dotfill\pageref{sec:supp_oneshot}
  \end{enumerate}
  \item[D.] Summary of notation\dotfill\pageref{sec:app_notation}
\end{enumerate}

\clearpage

\section{Implementation and Training Details}\label{sec:app_impl}

This section documents the complete architectural, optimization, data sampling, and baseline evaluation specifications required to reproduce \methodfour, \methodtwentyseven, and all evaluations reported on \bench.

\subsection{Backbone Architectures and Projection Head Configuration}\label{sec:supp_arch}

\textbf{Vision--language backbones.}
\methodfour is initialized from Qwen3-VL-4B~\cite{qwen3vl}, comprising a pretrained dynamic resolution vision transformer and a 36 layer causal language model with hidden dimension $D = 2{,}560$. During training, all vision encoder parameters remain frozen while all language model self-attention, feedforward, and normalization weights are updated in bfloat16 mixed precision with float32 master weights, scaled dot product attention, and activation gradient checkpointing. \methodtwentyseven scales to Qwen3.8-27B~\cite{qwen38} (hidden dimension $D = 5{,}120$), which interleaves 48 gated linear attention blocks (accelerated via Flash Linear Attention kernels) with 16 full causal self-attention blocks in a $3{:}1$ repeating layout across 64 transformer layers. To adapt visual correspondence across frames while preserving world knowledge and bounding memory overhead, we freeze the entire 27B base model and attach rank $256$ LoRA~\cite{LoRA} adapters ($\alpha = 64$, scaling factor $\alpha / r = 0.25$, dropout $0.05$, no bias) to the query, key, value, and output projections ($W_Q, W_K, W_V, W_O$) across the 16 full attention layers, yielding $167{,}772{,}160$ trainable LoRA parameters.

\textbf{Bilinear scale projection head.}
Both backbones attach an identical variable cardinality readout head (projection dimension $d = 256$) on top of the final transformer hidden states $H \in \mathbb{R}^{L \times D}$. Separate anchor and query projectors apply LayerNorm, a linear projection with bias $\mathbb{R}^{D} \to \mathbb{R}^{256}$, and GELU activation (Eq.~\ref{eq:proj}). For each active anchor $i \in \{1, \dots, K\}$, the concatenated interaction vector $\bm{f}_i = [\bm{a}_i \| \bm{q} \| (\bm{a}_i \odot \bm{q}) \| p_i] \in \mathbb{R}^{769}$ is scored by a shared two layer MLP mapping $\mathbb{R}^{769} \to \mathbb{R}^{256}$ (GELU) $\to \mathbb{R}^{1}$ (Eq.~\ref{eq:scorer}). This head introduces $1{,}518{,}849$ trainable parameters for \methodfour ($D = 2{,}560$) and $2{,}839{,}809$ trainable parameters for \methodtwentyseven ($D = 5{,}120$). Padded anchor positions $i > K$ (up to $K_{\max} = 11$) are masked prior to softmax normalization and soft cross-entropy computation. Architectural parameters are summarized in Appendix Table~\ref{tab:hp_model}.

\begin{table}[!htbp]
\centering
\caption{\textbf{Model architecture and adaptation hyperparameters for \methodfour and \methodtwentyseven.}}
\label{tab:hp_model}
\footnotesize
\renewcommand{\arraystretch}{1.12}
\resizebox{\linewidth}{!}{%
\begin{tabular}{lll}
\toprule
\textbf{Component / Parameter} & \textbf{\methodfour} & \textbf{\methodtwentyseven} \\
\midrule
Base VLM checkpoint & Qwen3-VL-4B~\cite{qwen3vl} & Qwen3.8-27B~\cite{qwen38} \\
Transformer layers / architecture & 36 full self-attention layers & 64 hybrid layers (48 linear + 16 full attn, $3{:}1$) \\
Backbone hidden dimension $D$ & 2{,}560 & 5{,}120 \\
Vision encoder status & Frozen & Frozen \\
Language backbone adaptation & Full parameter fine-tuning & Frozen base + rank 256 LoRA on 16 full attn layers \\
LoRA rank $r$ / alpha $\alpha$ / dropout & Not applicable & $r = 256$ / $\alpha = 64$ ($\alpha/r = 0.25$) / $0.05$ \\
LoRA target modules & Not applicable & Attention projections ($W_Q, W_K, W_V, W_O$) \\
Attention kernels & Scaled dot product attention & Scaled dot product + Flash Linear Attention \\
Head projection dimension $d$ & 256 & 256 \\
Anchor \& query projectors $\phi_{\mathrm{anc}}, \phi_{\mathrm{qry}}$ & LayerNorm $\to$ Linear($2560, 256$) $\to$ GELU & LayerNorm $\to$ Linear($5120, 256$) $\to$ GELU \\
Bilinear scale scorer MLP & Linear($769, 256$) $\to$ GELU $\to$ Linear($256, 1$) & Linear($769, 256$) $\to$ GELU $\to$ Linear($256, 1$) \\
Trainable LoRA / head parameters & -- / 1.52M (full LM + head tuned) & 167.77M LoRA + 2.84M head (170.61M total) \\
Compute / master weight precision & bfloat16 / float32 & bfloat16 / float32 \\
\bottomrule
\end{tabular}%
}
\end{table}

\subsection{Optimization, Data Mixture, and Dynamic Sampling Protocol}\label{sec:supp_train}

\textbf{Optimization schedule.}
Both models are optimized using AdamW with learning rate $1.0 \times 10^{-5}$, weight decay $0.01$, linear warmup over the first $10\%$ of total updates followed by cosine decay to zero, and global gradient norm clipping at $1.0$ with gradient checkpointing enabled. Training uses a batch size of 16 packed multimodal sequences per GPU without gradient accumulation and 10 asynchronous CPU data loading workers that perform trajectory selection on metadata before decoding only the selected reference and query frames. \methodfour is trained for up to 15{,}000 steps (checkpoints saved every 3{,}000 steps; release checkpoint at step 3{,}000), and \methodtwentyseven is trained for 2{,}500 steps (checkpoints saved every 500 steps; release checkpoint at step 1{,}500). Both runs share the exact same deterministic $1\%$ validation holdout per task.

\textbf{Balanced source dynamic sampling.}
The curated semantic training corpus totals 16{,}321 verified trajectories across eight data sources (Galaxea Open World: 7{,}574 across five R1-Lite partitions; RH20T: 4{,}873; LIBERO: 2{,}260; RoboArena: 745; AutoEval: 400; FINO-Net: 225; MetaWorld: 144; AgiBotWorld: 100), while all OOD evaluation benchmarks as well as external reward model training splits (e.g., RoboReward) are strictly excluded. At each training draw, one of the eight sources is chosen uniformly, a monotonic reference anchor set of initial cardinality $K_0 \sim \mathcal{U}\{3, \dots, 11\}$ is drawn from a successful or suboptimal trajectory and deduplicated across plateaus at threshold $\delta_{\mathrm{plat}} = 0.01$ by keeping one uniformly chosen frame per plateau, and up to 5 query search rounds (8 candidates per round) sample a trajectory of the same task across successful, suboptimal, and failed outcomes whose frame semantic progress lies within $[\tilde{a}, \tilde{b}]$. Training and sampling hyperparameters are listed in Appendix Table~\ref{tab:hp_sampling}.

\begin{table}[!htbp]
\centering
\caption{\textbf{Training optimization and dynamic in-context sampling hyperparameters.}}
\label{tab:hp_sampling}
\footnotesize
\renewcommand{\arraystretch}{1.12}
\begin{tabular}{ll}
\toprule
\textbf{Parameter} & \textbf{Value} \\
\midrule
Curated semantic training corpus & 16{,}321 trajectories across 8 manipulation sources \\
Validation split & Deterministic $1\%$ trajectory holdout per task \\
Source sampling strategy & Uniform across the 8 enabled datasets \\
Initial keyframe draw $K_0$ / surviving $K$ & $\mathcal{U}\{3, \dots, 11\}$ / $K \in \{2, \dots, 11\}$ \\
Reference outcome eligibility & Successful or suboptimal (nondecreasing before deduplication; strictly increasing after) \\
Plateau deduplication threshold $\delta_{\mathrm{plat}}$ / rule & $0.01$ / uniform single frame retention per plateau \\
Query outcome eligibility & Successful, suboptimal, and failed (same task \& split) \\
Candidate search budget & Up to 100 reference rounds; $5 \times 8$ query candidates per draw \\
Target formulation & Piecewise linear soft target in semantic progress space (Eq.~\ref{eq:soft_target}) \\
Optimizer / learning rate / weight decay & AdamW / $1.0 \times 10^{-5}$ / $0.01$ \\
LR schedule / warmup ratio / max grad norm & Cosine decay / $0.10$ / $1.0$ \\
Batch size per GPU / data workers / random seed & 16 / 10 / 42 \\
Total steps / save interval (\methodfour) & 15{,}000 / every 3{,}000 steps (selected step: 3{,}000) \\
Total steps / save interval (\methodtwentyseven) & 2{,}500 / every 500 steps (selected step: 1{,}500) \\
\bottomrule
\end{tabular}
\end{table}

\subsection{Baseline Evaluation Protocol on \bench}\label{sec:supp_baselines}

All 17 model configurations in Table~\ref{tab:main_benchmark} are evaluated on the exact same frozen \bench test split, containing 90 unseen tasks, 1{,}101 trajectories, 31{,}088 verified frames, 35{,}232 within trajectory pairs (32 randomly sampled pairs per trajectory), and 23{,}040 cross trajectory pairs (256 randomly sampled pairs per task reference group), where each sampled pair is randomly swapped with $50\%$ probability to balance ordering directions. For context conditioning, the $>10\text{T}$-parameter proprietary model \textbf{GPT-6-Astra} receives the identical interleaved reference keyframe set and anchor progress schedule per task group as \method; \textbf{GRM 2.0 Preview} (4B/8B) and \textbf{GRM} (3B/8B)~\cite{tan2025robodopamine}, which accept trajectory boundary context, receive the initial and terminal frames from the same reference trajectory; whereas text-driven open baselines---including Flora Rewarder~\cite{leng2026flora}, Robometer-4B (official 8 frame uniform window interpolated to 32 frames, as well as dense incremental prefix readout)~\cite{robometer2026}, RoboReward (4B/8B)~\cite{lee2026roboreward}, NimCritic~\cite{xu2026nimcritic}, DenseReward (single frame and 3 frame thinking)~\cite{fang2026densereward}, and RynnValue (4B/8B)~\cite{huang2026rynnvalue}---do not accept reference keyframes and are executed using their official released checkpoints, preprocessing pipelines, and native language-conditioned inference conventions mapped linearly onto $[0, 100]\%$. Because RynnValue predicts remaining time to completion rather than calibrated absolute task completion on $[0, 1]$, Terminal Agreement and direct MAE are undefined for RynnValue; following the benchmark specification, its pairwise rankings are evaluated by inverting remaining time predictions and its summary score is computed over Within and Cross Agreement ($^*$ in Table~\ref{tab:main_benchmark} and Appendix Table~\ref{tab:subset_breakdown}).

\section{Interleaved Multimodal Prompt Specification and Token Indexing}\label{sec:app_prompt}

Both during training and at inference time, \method formats a single turn multimodal user prompt without system prompt overhead or autoregressive text decoding. Each embedded image is enclosed by the backbone's native opening and closing visual boundary tokens. Appendix Box~B.1 displays the exact prompt structure formatted for a task instruction $l$ with $K$ reference keyframes $\{(k_i, p_i)\}_{i=1}^{K}$ and query observation $x_t$.

\begin{tcolorbox}[colback=metabg, colframe=metafg!35, boxrule=0.6pt, arc=4pt, left=8pt, right=8pt, top=6pt, bottom=6pt]
\small
\textbf{Box B.1: Interleaved multimodal prompt template used by \method.}\\[4pt]
{\ttfamily
Task: $\langle$task instruction $l\rangle$\\
The following ordered reference keyframes define relative progress for this query. The first reference means 0\% and the last means 100\% of this reference segment, not necessarily completion of the entire task.\\
BEGIN REFERENCE KEYFRAMES\\
Reference 1 of $K$, progress 0.0000\%:\\
$\langle\text{vision\_start}\rangle$ [Reference Keyframe Image $k_1$] $\langle\text{vision\_end}\rangle$}\hfill{\footnotesize$\leftarrow$ extract $\bm{h}_1 = H[\zeta_1]$}\\
{\ttfamily
Reference 2 of $K$, progress $\langle 100 \cdot p_2\rangle$\%:\\
$\langle\text{vision\_start}\rangle$ [Reference Keyframe Image $k_2$] $\langle\text{vision\_end}\rangle$}\hfill{\footnotesize$\leftarrow$ extract $\bm{h}_2 = H[\zeta_2]$}\\
$\vdots$\\
{\ttfamily
Reference $K$ of $K$, progress 100.0000\%:\\
$\langle\text{vision\_start}\rangle$ [Reference Keyframe Image $k_K$] $\langle\text{vision\_end}\rangle$}\hfill{\footnotesize$\leftarrow$ extract $\bm{h}_K = H[\zeta_K]$}\\
{\ttfamily
END REFERENCE KEYFRAMES\\
CURRENT QUERY IMAGE (not a reference):\\
$\langle\text{vision\_start}\rangle$ [Current Query Image $x_t$] $\langle\text{vision\_end}\rangle$\\
Locate the current state relative to the reference keyframes. Relative progress:}\hfill{\footnotesize$\leftarrow$ extract $\bm{h}_q = H[\zeta_{\mathrm{qry}}]$}
\end{tcolorbox}

Because sequences are padded on the right under causal masking, each formatted prompt contains exactly $K + 1$ closing image boundary tokens: the first $K$ boundary positions define the anchor extraction indices $(\zeta_1, \dots, \zeta_K)$ yielding $(\bm{h}_1, \dots, \bm{h}_K)$, and the final valid token index $\zeta_{\mathrm{qry}}$ provides the query hidden state $\bm{h}_q$ that attends across the entire multimodal sequence.

\section{Extended Benchmark Breakdown, Checkpoint Sweep, and Domain Generalization}\label{sec:app_extra_eval}

\subsection{Subset Evaluation on \bench}\label{sec:supp_subsets}

To diagnose model behavior across distinct deployment conditions, \bench partitions its 1{,}101 OOD trajectories into three complementary diagnostic subsets:
(1)~\textbf{Real World OOD} (502 full length trajectories across 40 unseen real robot tasks, 16{,}064 frames), testing general calibration across diverse objects and scenes;
(2)~\textbf{Truncated Prefixes} (259 incomplete 16 frame trajectory prefixes clipped midway, 4{,}144 frames), explicitly testing whether a model hallucinates 100\% completion at the end of an arbitrarily clipped video; and
(3)~\textbf{RoboFAC Failure Analysis Subset} (340 trajectories across 50 unseen tasks from RoboFAC~\cite{lu2025robofac}, 10{,}880 frames), stressing detailed discrimination between successful completions and diverse manipulation failure modes.

Appendix Table~\ref{tab:subset_breakdown} reports Average, Within Agreement, Cross Agreement, Terminal Agreement, and MAE across all three subsets for all 17 evaluated models. Two sharp patterns stand out. First, on \textbf{Truncated Prefixes}, conventional video clip reward models suffer catastrophic terminal miscalibration because they implicitly equate the final frame of a video buffer with episode completion: Robometer-4B drops to 37.1\% Terminal Agreement (58.0\% Average, 19.5 pp MAE) and Flora Rewarder drops to 40.5\% Terminal Agreement (57.9\% Average), whereas \methodtwentyseven achieves \textbf{81.1\%} Terminal Agreement, \textbf{74.7\%} Average, and \textbf{7.4} pp MAE, even surpassing GPT-6-Astra (74.0\% Average, 78.8\% Terminal Agreement, 8.9 pp MAE). Second, on \textbf{RoboFAC}, \methodtwentyseven attains \textbf{77.5\%} Average and \textbf{9.6} pp MAE, outperforming Robometer-4B by +13.9 Average points ($-17.7$ pp MAE) and GPT-6-Astra by +1.2 Average points ($-1.6$ pp MAE).

\begin{table}[!htbp]
\centering
\caption{\textbf{Subset evaluation breakdown across all 17 models on \bench.}
We report Average (\textbf{Avg}), Within Agreement (\textbf{Win}), Cross Agreement (\textbf{Cross}), Terminal Agreement (\textbf{Term}), and frame \textbf{MAE} ($\downarrow$) on Real World OOD (502 trajectories), Truncated Prefixes (259 trajectories), and RoboFAC (340 trajectories). Best overall results per column are in \textbf{bold}; runner up results are \underline{underlined}. $^*$ denotes remaining time models where Terminal Agreement and MAE are not applicable.}
\label{tab:subset_breakdown}
\footnotesize
\setlength{\tabcolsep}{3.1pt}
\renewcommand{\arraystretch}{1.10}
\resizebox{\textwidth}{!}{%
\begin{tabular}{@{}lccccc ccccc ccccc@{}}
\toprule
& \multicolumn{5}{c}{\textbf{Real World OOD (502 traj, 40 tasks)}} & \multicolumn{5}{c}{\textbf{Truncated Prefixes (259 traj)}} & \multicolumn{5}{c}{\textbf{RoboFAC Subset (340 traj, 50 tasks)}} \\
\cmidrule(lr){2-6}\cmidrule(lr){7-11}\cmidrule(lr){12-16}
\textbf{Model} & \textbf{Avg}$\uparrow$ & \textbf{Win}$\uparrow$ & \textbf{Cross}$\uparrow$ & \textbf{Term}$\uparrow$ & \textbf{MAE}$\downarrow$ & \textbf{Avg}$\uparrow$ & \textbf{Win}$\uparrow$ & \textbf{Cross}$\uparrow$ & \textbf{Term}$\uparrow$ & \textbf{MAE}$\downarrow$ & \textbf{Avg}$\uparrow$ & \textbf{Win}$\uparrow$ & \textbf{Cross}$\uparrow$ & \textbf{Term}$\uparrow$ & \textbf{MAE}$\downarrow$ \\
\midrule
\rowcolor{yuhengBg}\methodtwentyseven (Ours) & \underline{74.4} & \textbf{73.5} & \underline{75.1} & 74.7 & \textbf{10.6} & \textbf{74.7} & \underline{72.8} & 70.2 & \textbf{81.1} & \textbf{7.4} & \textbf{77.5} & \textbf{75.8} & \textbf{73.8} & \textbf{82.9} & \textbf{9.6} \\
GPT-6-Astra (Proprietary) & \textbf{75.0} & 71.4 & 72.4 & \textbf{81.3} & 13.4 & \underline{74.0} & 72.5 & \underline{70.6} & \underline{78.8} & 8.9 & \underline{76.3} & \underline{74.6} & \underline{72.2} & \underline{82.1} & \underline{11.2} \\
\rowcolor{yuhengBg}\methodfour (Ours) & 73.4 & \underline{72.7} & \textbf{76.1} & 71.3 & \underline{12.0} & 72.5 & \textbf{73.8} & \textbf{71.6} & 72.2 & \underline{8.0} & 74.3 & 73.0 & 70.2 & 79.7 & 11.4 \\
\midrule
GRM 2.0 Preview 8B~\cite{tan2025robodopamine} & 64.8 & 63.9 & 54.5 & \underline{76.1} & 21.7 & 64.1 & 67.7 & 55.8 & 68.7 & 19.2 & 64.6 & 63.9 & 57.8 & 72.1 & 23.2 \\
Flora Rewarder~\cite{leng2026flora} & 67.7 & 68.7 & 69.3 & 65.1 & 19.8 & 57.9 & 67.1 & 66.0 & 40.5 & 13.1 & 63.2 & 67.5 & 65.6 & 56.5 & 20.4 \\
Robometer-4B (Global 8 frame interp.)~\cite{robometer2026} & 66.3 & 70.4 & 67.0 & 61.4 & 20.9 & 58.0 & 71.5 & 65.4 & 37.1 & 19.5 & 63.6 & 68.2 & 64.7 & 57.9 & 27.3 \\
GRM 2.0 Preview 4B~\cite{tan2025robodopamine} & 60.9 & 61.1 & 53.5 & 67.9 & 21.7 & 62.5 & 64.8 & 56.2 & 66.4 & 18.0 & 65.6 & 64.6 & 58.2 & 74.1 & 21.6 \\
RoboReward 8B~\cite{lee2026roboreward} & 63.5 & 64.9 & 67.3 & 58.2 & 18.4 & 66.6 & 69.6 & 66.9 & 63.3 & 9.6 & 56.2 & 60.3 & 56.8 & 51.5 & 22.1 \\
RoboReward 4B~\cite{lee2026roboreward} & 64.4 & 65.6 & 66.3 & 61.4 & 20.7 & 61.5 & 67.8 & 64.5 & 52.1 & 11.3 & 55.2 & 59.4 & 54.9 & 51.2 & 24.0 \\
Robometer-4B (Incremental prefix)~\cite{robometer2026} & 64.0 & 65.9 & 65.2 & 61.0 & 25.8 & 55.4 & 68.4 & 64.4 & 33.6 & 22.4 & 57.9 & 60.0 & 56.0 & 57.6 & 37.1 \\
NimCritic~\cite{xu2026nimcritic} & 59.6 & 65.4 & 57.2 & 56.2 & 26.9 & 45.5 & 64.5 & 54.5 & 17.4 & 22.5 & 46.4 & 53.1 & 45.0 & 41.2 & 41.3 \\
DenseReward (single frame)~\cite{fang2026densereward} & 47.7 & 49.7 & 48.9 & 44.4 & 29.2 & 50.6 & 52.0 & 52.6 & 47.1 & 23.1 & 46.3 & 49.1 & 47.0 & 42.6 & 27.8 \\
DenseReward (3 frame thinking)~\cite{fang2026densereward} & 48.3 & 51.5 & 51.4 & 41.8 & 28.4 & 50.4 & 52.7 & 51.3 & 47.1 & 22.8 & 44.3 & 50.6 & 50.1 & 32.4 & 27.6 \\
GRM 8B~\cite{tan2025robodopamine} & 41.1 & 45.6 & 38.2 & 39.4 & 25.5 & 58.9 & 58.9 & 52.5 & 65.3 & 13.2 & 48.1 & 49.0 & 39.6 & 55.6 & 26.1 \\
GRM 3B~\cite{tan2025robodopamine} & 40.2 & 45.2 & 34.0 & 41.4 & 27.6 & 58.9 & 59.1 & 50.7 & 66.8 & 13.6 & 45.0 & 47.0 & 36.0 & 52.1 & 25.5 \\
RynnValue 8B$^*$~\cite{huang2026rynnvalue} & 43.6 & 41.2 & 45.9 & -- & -- & 27.9 & 25.1 & 30.7 & -- & -- & 42.8 & 39.5 & 46.2 & -- & -- \\
RynnValue 4B$^*$~\cite{huang2026rynnvalue} & 37.6 & 34.5 & 40.8 & -- & -- & 20.9 & 19.6 & 22.3 & -- & -- & 35.3 & 31.3 & 39.2 & -- & -- \\
\bottomrule
\end{tabular}%
}
\end{table}

\subsection{Complete Training Checkpoint Sweep}\label{sec:supp_ckpts}

Appendix Table~\ref{tab:ckpt_sweep} reports the complete evaluation metrics on \bench across all saved checkpoints of \methodtwentyseven (steps 500--2{,}500) and \methodfour (steps 3{,}000--15{,}000), complementing the training trajectory curves in Fig.~\ref{fig:ablations}C. The sweep confirms the robustness of the released checkpoints (step 1{,}500 for \methodtwentyseven and step 3{,}000 for \methodfour, fixed prior to OOD evaluation): \methodtwentyseven remains tightly clustered across steps 500--2{,}500 (74.8\%--75.7\% Average, 9.8--11.0 pp MAE), while full-LM \methodfour converges rapidly by step 3{,}000 (73.1\% Within, 11.3 pp MAE) with only mild fluctuations on extended updates (step 9{,}000 reaches 73.6\% Average via Terminal agreement while fine-grained Within and MAE edge to 72.6\% and 11.4 pp). Across every saved checkpoint of both models, Average agreement remains between 71.8\% and 75.7\% and MAE remains between 9.8 and 12.0 pp, outperforming the strongest open baseline (GRM 2.0 Preview 8B, 64.9\% Average, 21.9 pp MAE) by 6.9 to 10.8 pp regardless of checkpoint selection.

\begin{table}[!htbp]
\centering
\caption{\textbf{Complete checkpoint sweep for \methodtwentyseven and \methodfour on \bench (1{,}101 trajectories).}
Shaded rows indicate the released model checkpoints evaluated throughout the main text (step 1{,}500 for \methodtwentyseven; step 3{,}000 for \methodfour). Best values per column are in \textbf{bold}.}
\label{tab:ckpt_sweep}
\footnotesize
\setlength{\tabcolsep}{3.8pt}
\renewcommand{\arraystretch}{1.12}
\resizebox{\textwidth}{!}{%
\begin{tabular}{@{}lc c ccc cccc c ccc@{}}
\toprule
& & & \multicolumn{3}{c}{\textbf{Within Agreement (\%) $\uparrow$}} & \multicolumn{4}{c}{\textbf{Cross Agreement (\%) $\uparrow$}} & & \multicolumn{3}{c}{\textbf{Progress MAE (pp) $\downarrow$}} \\
\cmidrule(lr){4-6}\cmidrule(lr){7-10}\cmidrule(lr){12-14}
\textbf{Model Variant} & \textbf{Step} & \textbf{Avg $\uparrow$} & \textbf{All} & \textbf{Succ.} & \textbf{Unsucc.} & \textbf{All} & $\mathbf{S{\leftrightarrow}S}$ & $\mathbf{N{\leftrightarrow}N}$ & $\mathbf{S{\leftrightarrow}N}$ & \textbf{Term. $\uparrow$} & \textbf{All} & \textbf{Succ.} & \textbf{Unsucc.} \\
\midrule
\methodtwentyseven & 500 & 75.0 & 72.6 & 79.8 & 67.8 & 73.2 & 78.3 & 65.5 & 76.6 & 79.1 & 10.9 & 10.3 & 11.4 \\
\methodtwentyseven & 1{,}000 & 74.8 & 72.3 & 80.2 & 67.0 & 73.5 & \textbf{79.0} & 66.2 & 76.6 & 78.5 & 11.0 & 9.8 & 11.9 \\
\rowcolor{yuhengBg}\methodtwentyseven (Release) & 1{,}500 & \textbf{75.7} & \textbf{74.0} & 80.9 & \textbf{69.4} & \textbf{74.2} & \textbf{79.0} & 67.3 & \textbf{77.3} & 78.7 & \textbf{9.8} & 9.3 & \textbf{10.2} \\
\methodtwentyseven & 2{,}000 & 75.4 & 73.1 & 80.7 & 68.1 & 73.6 & 78.8 & 66.3 & 76.8 & \textbf{79.5} & 10.2 & \textbf{9.2} & 11.0 \\
\methodtwentyseven & 2{,}500 & 75.6 & 73.7 & \textbf{81.1} & 68.8 & 73.9 & \textbf{79.0} & 66.6 & 77.2 & 79.3 & \textbf{9.8} & \textbf{9.2} & 10.4 \\
\midrule
\rowcolor{yuhengBg}\methodfour (Release) & 3{,}000 & 73.3 & 73.1 & 78.7 & 69.3 & 72.8 & 77.9 & 66.9 & 75.1 & 74.1 & 11.3 & 11.1 & 11.4 \\
\methodfour & 6{,}000 & 72.5 & 71.7 & 78.6 & 67.1 & 72.2 & 76.9 & 66.4 & 74.6 & 73.7 & 12.0 & 11.0 & 12.7 \\
\methodfour & 9{,}000 & 73.6 & 72.6 & 78.5 & 68.6 & 73.2 & 77.3 & \textbf{68.4} & 75.2 & 74.9 & 11.4 & 10.7 & 12.0 \\
\methodfour & 12{,}000 & 72.3 & 71.4 & 76.8 & 67.8 & 72.4 & 75.4 & \textbf{68.4} & 74.1 & 73.1 & 12.0 & 11.7 & 12.3 \\
\methodfour & 15{,}000 & 71.8 & 71.1 & 76.3 & 67.6 & 72.3 & 75.2 & 68.1 & 74.1 & 71.9 & 12.0 & 11.8 & 12.1 \\
\bottomrule
\end{tabular}%
}
\end{table}

\subsection{One-Shot Generalization Across Manipulation Domains}\label{sec:supp_oneshot}

\begin{figure}[!p]
\centering
\includegraphics[width=\textwidth,height=0.86\textheight,keepaspectratio]{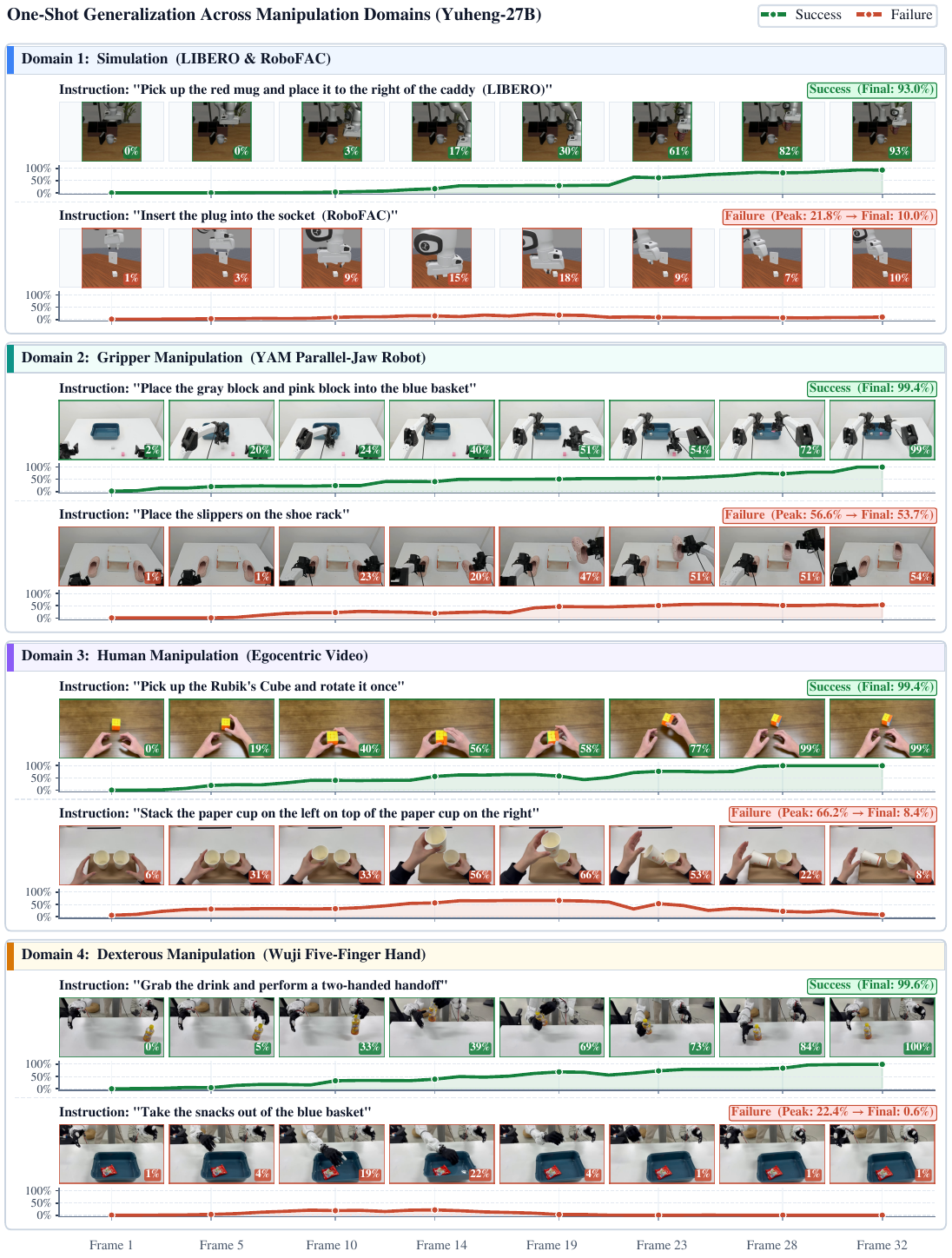}
\vspace{-4pt}
\caption{\textbf{One-shot generalization across four manipulation domains.}
Evaluated with \methodtwentyseven conditioned on a single successful demonstration per unseen task (raw unsmoothed frame predictions, no endpoint clamping). Trajectories are organized in pairs (one Success and one Failure) across four domain groups:
(\textbf{1})~\emph{Simulation (LIBERO \& RoboFAC)},
(\textbf{2})~\emph{Gripper Manipulation (YAM Parallel-Jaw Robot)},
(\textbf{3})~\emph{Human Manipulation (Egocentric Video)}, and
(\textbf{4})~\emph{Dexterous Manipulation (Wuji Five-Finger Hand)}.
Each trajectory displays the task instruction, eight uniformly sampled frames (frames 1, 5, 10, 14, 19, 23, 28, and 32) with per-frame predicted progress annotations, and an aligned full-width progress curve underneath.}
\label{fig:app_oneshot}
\end{figure}

To inspect per-frame progress predictions beyond aggregate benchmark statistics, Appendix Fig.~\ref{fig:app_oneshot} visualizes eight unseen trajectories spanning four distinct manipulation domains---Simulation (LIBERO and RoboFAC), Gripper Manipulation (YAM parallel jaw robot), Human Manipulation (egocentric video), and Dexterous Manipulation (Wuji five finger hand)---with two trajectories (one successful and one failed) per domain. Each unseen task is evaluated by \methodtwentyseven conditioned solely on a single successful reference demonstration, reporting raw unsmoothed frame scores without endpoint clamping. Aligning eight uniformly sampled frames directly above each full-width progress curve illustrates how \methodtwentyseven tracks milestone transitions on successful rollouts (reaching 93.0\%--99.6\%) while stalling at partial progress or regressing sharply upon failed grasps, missed insertions, or dropped objects (ending at 0.6\%--53.7\%).

\section{Summary of Notation}\label{sec:app_notation}

Appendix Table~\ref{tab:notation} summarizes the mathematical notation used throughout the main text and appendices.

\begin{table}[!htbp]
\caption{\textbf{Summary of notation used throughout this technical report.}}
\label{tab:notation}
\centering
\footnotesize
\renewcommand{\arraystretch}{1.10}
\begin{tabular}{@{}ll@{}}
\toprule
\textbf{Symbol} & \textbf{Description} \\
\midrule
\multicolumn{2}{@{}l}{\textit{Tasks, trajectories, and reference scales}} \\
$l$ & Natural language manipulation task instruction \\
$\tau^{\mathrm{ref}} = (o^{\mathrm{ref}}_1, \dots, o^{\mathrm{ref}}_{T_{\mathrm{ref}}})$ & Successful or monotonic reference demonstration of length $T_{\mathrm{ref}}$ \\
$\tau = (x_1, \dots, x_T)$ & Arbitrary query trajectory of length $T$ (successful, suboptimal, or failed) \\
$c(\tau) \in \{\text{success}, \text{suboptimal}, \text{failure}\}$ & Verified semantic trajectory outcome label \\
$K$ ($K_0$, $K_{\max}$) & Active reference keyframe count ($K \ge 2$); initial sample count $K_0$; batch cap $K_{\max} = 11$ \\
$\mathcal{K} = (k_1, \dots, k_K)$ & Ordered visual reference keyframes extracted from $\tau^{\mathrm{ref}}$ \\
$\mathcal{P} = (p_1, \dots, p_K)$ & Strictly increasing normalized anchor progress scale satisfying $0 = p_1 < \dots < p_K = 1$ \\
\midrule
\multicolumn{2}{@{}l}{\textit{Multimodal representations and bilinear scale head}} \\
$F_\theta$ & Universal in-context progress model with trainable parameters $\theta$ \\
$H \in \mathbb{R}^{L \times D}$ & Final layer transformer hidden states of sequence length $L$ and hidden width $D$ \\
$\zeta_i$, $\zeta_{\mathrm{qry}}$ & Token indices of the $i$-th closing image boundary token and terminal query token \\
$\bm{h}_i, \bm{h}_q \in \mathbb{R}^{D}$ & Extracted hidden vectors for anchor $i$ and query observation $x_t$ \\
$\phi_{\mathrm{anc}}, \phi_{\mathrm{qry}}$ & LayerNorm--Linear--GELU projection modules into latent alignment dimension $d = 256$ \\
$\bm{a}_i, \bm{q} \in \mathbb{R}^{d}$ & Projected latent embeddings of reference anchor $i$ and query frame $x_t$ \\
$\bm{f}_i \in \mathbb{R}^{3d + 1}$ & Bilinear scale interaction vector $[\bm{a}_i \| \bm{q} \| (\bm{a}_i \odot \bm{q}) \| p_i]$ \\
$z_i \in \mathbb{R}$, $\pi_i \in [0, 1]$ & Anchor compatibility logit and masked softmax probability over $i \in \{1, \dots, K\}$ \\
$\hat{p}_t = \sum_{i=1}^K \pi_i p_i \in [0, 1]$ & Predicted continuous relative semantic progress for query frame $x_t$ \\
\midrule
\multicolumn{2}{@{}l}{\textit{Semantic space sampling and training objective}} \\
$\tilde{p}_t, \tilde{p}^{\star} \in [0, 1]$ & Calibrated semantic frame progress on reference and query trajectories \\
$\delta_{\mathrm{plat}}$ & Plateau deduplication threshold ($\delta_{\mathrm{plat}} = 0.01$) for collapsing near ties \\
$[\tilde{a}, \tilde{b}]$ & Raw semantic progress span $[\tilde{p}_{t_1}, \tilde{p}_{t_K}]$ of surviving reference keyframes \\
$p^{\star} \in [0, 1]$ & Segment normalized ground truth semantic progress $(\tilde{p}^{\star} - \tilde{a}) / (\tilde{b} - \tilde{a})$ \\
$\alpha \in [0, 1]$ & Piecewise linear interpolation ratio $(p^{\star} - p_\ell) / (p_{\ell+1} - p_\ell)$ between bracketing anchors \\
$\bm{y} = (y_1, \dots, y_K) \in \Delta^{K-1}$ & Soft classification target distribution over two adjacent anchors in progress space \\
$\mathcal{L}_{\mathrm{prog}}(\theta)$ & Soft cross-entropy training loss $-\sum_{i=1}^{K} y_i \log \pi_i$ \\
\bottomrule
\end{tabular}
\end{table}

\end{appendices}